\pdfoutput=1
\documentclass[11pt]{article}

\usepackage[final]{acl}

\usepackage{times}
\usepackage{latexsym}
\usepackage{adjustbox}
\usepackage{longtable}
\usepackage{multirow}

\usepackage[T1]{fontenc}
\usepackage[utf8]{inputenc}
\usepackage{subcaption}
\usepackage{placeins}

\usepackage{microtype}

\usepackage{inconsolata}
\usepackage{float}
\usepackage{caption}
\usepackage{amssymb}
\usepackage{graphicx}
\usepackage{booktabs}
\usepackage{adjustbox}

\usepackage{footmisc}
\usepackage{amsmath}

\usepackage{xcolor}
\usepackage{enumitem}

\usepackage{multirow}
\usepackage{array}
\usepackage{longtable}
\usepackage{afterpage}

\usepackage{algorithm}
\usepackage{algpseudocode}
\usepackage{amsmath}
\usepackage{array}
\usepackage{graphicx} % Optional, for adjusting box heights
\usepackage{caption}
\usepackage{enumitem}
\usepackage{tcolorbox}

\title{\href{}{Towards Robust ESG Analysis Against Greenwashing Risks}: \\Aspect-Action Analysis with Cross-Category Generalization}
\author{
    Keane Ong\textsuperscript{$\spadesuit$}, 
    Rui Mao\textsuperscript{$\clubsuit$}, 
    Deeksha Varshney\textsuperscript{$\spadesuit$},
    \bf Erik Cambria\textsuperscript{$\clubsuit$}, 
    Gianmarco Mengaldo\textsuperscript{$\spadesuit$}\thanks{Corresponding author} \\
    \textsuperscript{$\spadesuit$}National University of Singapore, Singapore \\
    \textsuperscript{$\clubsuit$}Nanyang Technological University, Singapore \\
    \texttt{\{keane.ongweiyang\}@u.nus.edu};
     \texttt{\{deeksha2,mpegim\}@nus.edu.sg};\\
     \texttt{\{rui.mao,cambria\}@ntu.edu.sg} 
}

\begin{document}
% make the title area
\maketitle
\begin{abstract}
Sustainability reports are key for evaluating companies’ environmental, social and governance (ESG) performance. To analyze these reports, NLP approaches can efficiently extract ESG insights at scale. However, even the most advanced NLP methods lack robustness against ESG content that is greenwashed – i.e. sustainability claims that are misleading, exaggerated, and fabricated. Accordingly, existing NLP approaches often extract insights that reflect misleading or exaggerated sustainability claims rather than objective ESG performance. To tackle this issue, we introduce A3CG - \textbf{A}spect-\textbf{A}ction \textbf{A}nalysis with Cross-\textbf{C}ategory \textbf{G}eneralization, as a novel dataset to improve the robustness of ESG analysis amid the prevalence of greenwashing. By explicitly linking sustainability aspects with their associated actions, A3CG facilitates a more fine-grained and transparent evaluation of sustainability claims, ensuring that insights are grounded in verifiable actions rather than vague or misleading rhetoric. Additionally, A3CG emphasizes cross-category generalization. This ensures robust model performance in aspect-action analysis even when companies change their reports to selectively favor certain sustainability areas. Through experiments on A3CG, we analyze state-of-the-art supervised models and LLMs, uncovering their limitations and outlining key directions for future research.

\end{abstract}

\section{Introduction}

Sustainability reports have become an important mechanism for evaluating a company's environmental, social, and governance (ESG) performance~\cite{nguyen2020empiricalsusreport}. To facilitate the analysis of these reports, NLP methods typically offer an automated and efficient way to extract ESG insights at scale~\cite{ong2024explainable}. However, the rise of greenwashing – the practice of providing misleading, exaggerated, and fabricated sustainability claims – has increasingly obscured the content of these reports~\cite{ong2024explainable}. Greenwashing presents a major challenge for NLP-driven ESG analysis, let alone manual assessments, ultimately hindering meaningful progress toward sustainability goals~\cite{rajesh2020susperfimpt}.

% sentiment analysis~\cite{song2018sentimentanalysissustainable}
Yet, existing NLP methods for ESG analysis - i.e. topic analysis~\cite{ong2025esgsenticnet}, retrieval-augmented generation (RAG)~\cite{esgrevealrag}, fail to account for greenwashing risks. These approaches aim to provide actionable insights from sustainability reports, but fail to account for the credibility of the claims within them. Therefore, the insights extracted by these NLP methods can often reflect vague and misleading rhetoric instead of genuine sustainability initiatives~\cite{ong2024explainable}.

\begin{figure}[t]
    \centering
    \includegraphics[width=\linewidth]{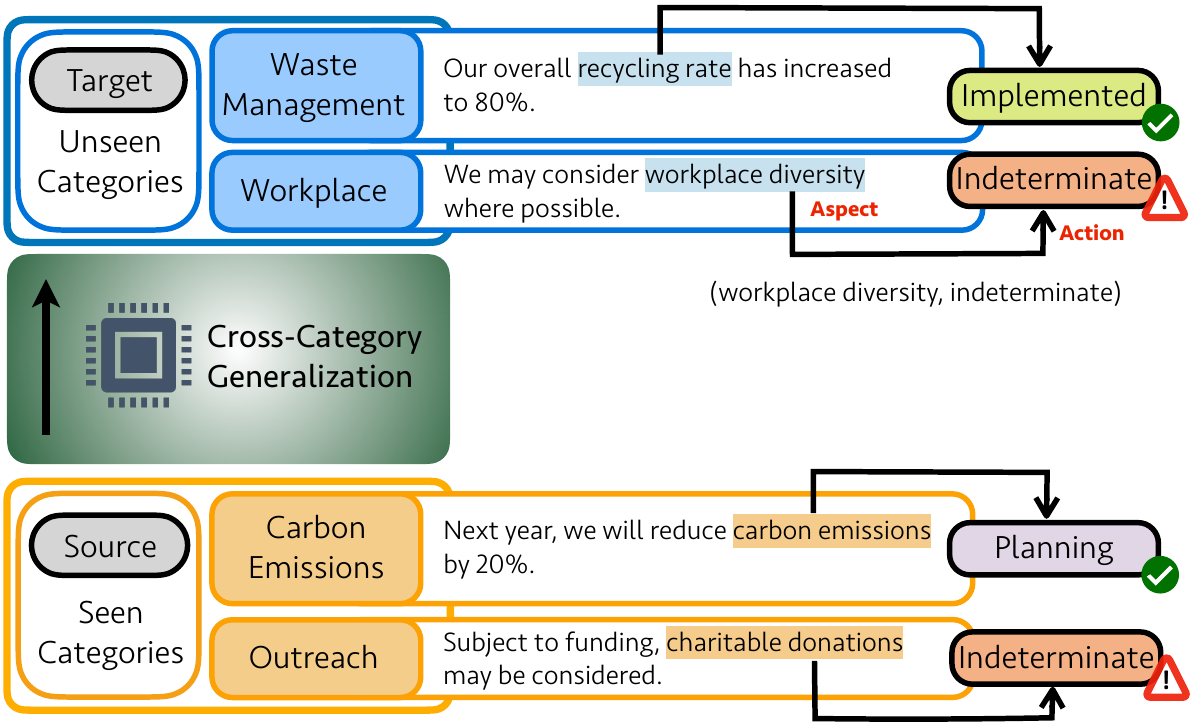}
    \caption{A3CG Task Overview. Given a sustainability statement, the model extracts aspect–action pairs. To evaluate cross-category generalization, it is tested on statements containing aspect categories not seen during training -- i.e. trained on "source", tested on "target".}
    \label{fig:overview_a3cg}
\end{figure}
To tackle this, we propose A3CG\footnote{We release the dataset and experimental codes at: \\ \url{https://github.com/keanepotato/a3cg_greenwash}} - \textbf{A}spect-\textbf{A}ction \textbf{A}nalysis with Cross-\textbf{C}ategory \textbf{G}eneralization, a novel dataset for robust ESG analysis amid the prevalence of greenwashing. A3CG provides a foundation for systematically extracting two interconnected components from sustainability texts: \textit{i) Aspect:} The sustainability-related entity, goal, or activity discussed (i.e. ``carbon emissions reduction''). \textit{ii) Action:} The type of engagement related to the aspect, ``implemented'', ``planning'', or ``indeterminate''. While ``implemented'' and ``planning'' indicate concrete steps taken or planned toward the sustainability aspect, ``indeterminate'' highlights vague, evasive or non-attributable claims which may be linked with greenwashing~\cite{siano2017greenwashingmarkers}. By enabling the identification of sustainability aspects and their action assessments, A3CG allows models to clarify genuine sustainability initiatives from ambiguous or misleading rhetoric. Through this flagging of clear sustainability initiatives, stakeholders can then verify and track them more easily. Therefore, while A3CG is not designed for \textit{definitive} greenwashing detection, it facilitates \textit{clear} and \textit{transparent} ESG analysis that is less susceptible to greenwashing risks.

To further enhance the robustness of ESG analysis, A3CG extends aspect-action analysis to a cross-category generalization setting. This ensures that aspect-action analysis remains effective even when companies change their report content to selectively favor certain sustainability areas~\cite{Darnall2022}. By emphasizing the extraction of aspect-action pairs from sustainability categories beyond those seen in training, A3CG facilitates aspect-action analysis on new, previously unencountered sustainability themes. As a result, aspect-action analysis remains robust in distinguishing genuine sustainability initiatives from rhetoric, even with changes in report content.

% This ensures the ESG analysis remains robust even in cases of selective reporting~\cite{ong2024explainable} - where companies strategically rebrand their sustainability reports by emphasizing specific favorable sustainability areas without necessarily improving sustainability performance~\cite{Darnall2022}.

% ~\cite{derqui2020towardsevolving}. 

We conduct extensive experiments on A3CG, evaluating the performance of state-of-the-art (SOTA) supervised models and large language models (LLMs). Our results highlight: (1) Supervised learning methods - GRACE~\cite{Luo2020grace}, outperform the latest LLMs - Claude 3.5 Sonnet~\cite{huang2024olympicarenaclaude3.5}, DeepSeek V3~\cite{liu2024deepseek}, on A3CG. (2) Contrastive learning is relatively more effective than adversarial learning for cross-category generalization. (3) The limitations of supervised models and LLMs in A3CG subtasks and linguistic cases. (4) Research directions and strategies for tackling these limitations.

Our main contributions include: (1) We develop and release A3CG, the first dataset tailored for ensuring robust ESG analysis amid the prevalence of greenwashing risks. (2) We conduct extensive experiments on A3CG using SOTA supervised learning and LLM methods, uncovering valuable insights for NLP research and model development.

\section{Related Work}

\noindent\textbf{NLP Methods in Sustainability Analysis.}
To automate the extraction of insights from sustainability reports, popular NLP approaches include analyzing topic occurrence~\cite{ong2025esgsenticnet}, sentiment analysis~\cite{song2018sentimentanalysissustainable}, and RAG~\cite{esgrevealrag}. Yet, these traditional NLP approaches do not consider how greenwashing might distort extracted insights~\cite{ong2024explainable}. To promote the transparency of sustainability claims,~\citet{stammbach2022environmental} studied environmental claim detection, and~\citet{ni2023chatreport} assessed report conformance to TCFD guidelines. However, these works do not explicitly address building robust ESG analysis against greenwashing risks.

% , zhou2021topicdiscovery
% pasch2022sentimentclass

\noindent\textbf{Aspect-Based Sentiment Analysis}.
Aspect-Based Sentiment Analysis (ABSA) has become a popular NLP problem over the past decade~\cite{duuinc,ong2023finxabsa}. Yet, ABSA has focused on reviews and opinion mining~\cite{heemet}, despite calls for aspect-level analysis to be extended to other contexts~\cite{chebolu2023review}. Our study presents the first adaptation of aspect-level analysis to the aspect-action classification task. While we focus on sustainability, aspect-action analysis could potentially be adapted for other applications - i.e. government policy analysis~\cite{howlett2015policyanalysis}. 
% financial report analysis~\cite{liu2016fin10k}. 

\noindent \textbf{Cross-Generalization Studies}. Cross-domain generalization has become an increasingly important challenge in NLP~\cite{wang2022generalizingstudy}. However, there has been limited investigation on how models perform on \textit{unseen categories within the same domain} - i.e. cross-category generalization. Yet, this deserves attention because intra-domain data shifts are distinct from cross-domain shifts, and can significantly degrade model performance~\cite{subbaswamy2021intradomainshift}. Moreover, data annotation for unseen categories is expensive and impractical, as these categories often arise unexpectedly from evolving text content~\cite{bai2024bvsp}. 

% \noindent\textbf{Cross-Generalization Methods.}
% Popular methods for cross-domain generalization include learning domain-invariant features through adversarial training~\cite{yuan2024advcrossdomain}, or learning higher-level semantic features through contrastive learning~\cite{xie2022contrastivecrossdomain}. Yet, it remains unclear whether these techniques can be effective for cross-category generalization.

\noindent \textbf{LLMs Biases in Sustainability Analysis.}
Given LLMs' reasoning capabilities~\cite{wei2022emergent}, they have increasingly been utilized for tasks within sustainability analysis, such as knowledge base construction~\cite{ong2025esgsenticnet}, RAG~\cite{esgrevealrag}, among others. However, their performance on domain-specific tasks can be impacted by pre-training biases~\cite{dai2024llmbiassurvey, yang2024mitigating}, and the impact of these biases on sustainability analysis tasks remains under-explored.

\section{Dataset Construction}
This section describes A3CG's construction.
%  are available in this anonymous 
% \href{https://drive.google.com/drive/folders/1LBBzQSmKSsmeABwWYWF19FUSOjr52XnL?usp=share_link}{link}.

\subsection{Data Collection \& Quality Control}
We collect sustainability statements from 1679 sustainability reports\footnote{Company names and the corresponding years of sustainability reports are also available at: \url{https://github.com/keanepotato/a3cg_greenwash}. Using this information, these public reports can be searched and accessed online.} of Singapore Exchange (SGX) companies, from 2017-2022 inclusive. The reports are public and can be found online via company websites. For quality control, we eliminate: (1) Incomplete statements (2) Incoherent statements with misspelled words (3) Non-English statements. 

\subsection{Annotation Scheme}

Pairs of (aspect, action) are annotated for each sustainability statement. In a pair, the \textit{action} characterizes the company's engagement with the \textit{aspect}. Notably, the aspect's sustainability category (i.e. "emissions control" as a category of the aspect "carbon emissions") is also annotated. However,  similar to other generalization studies~\cite{xu2023measuringaste}, the aspect category is not a prediction target. Instead, it is utilized to split the dataset for cross-category testing. We summarize the aspect and action definitions\footnote{Detailed annotation instructions, guidelines, definitions, samples, and annotator details can be found in appendix~\ref{sec:humanannotate}} below: \\
\noindent\textbf{Aspect:} A sustainability aspect is defined as a significant~\textit{entity, goal, sub-area, activity} of a sustainability category, providing a focal point for an action to address or engage with. We focus on aspects that  belong to the sustainability categories in Table~\ref{tab:taxonomy}, and are explicitly found within a statement. \\
\noindent\textbf{Action:} The type of action taken toward the aspect. \textit{i) Planning:} Indicates that an action has been planned or a commitment has been made by a company to address or engage with the \textit{aspect} to advance its sustainability efforts. \textit{ii) Implemented:} Indicates that an action has already been taken to address or engage with the \textit{aspect} to advance its sustainability efforts. \textit{iii) Indeterminate:} Indicates that it is unclear from the statement if the company intends to address or engage with the \textit{aspect} to advance its sustainability efforts, or how it intends to do so. 

For i) and ii), ``addressing'' or ``engaging with'' an aspect involves incorporating the aspect within company operations if it is a sustainability \textit{activity} or \textit{entity}, or \textit{advancing} the aspect if it is a sustainability \textit{goal or sub-area}. On the other hand, iii) characterizes vague, non-committal and non-attributable language. These action labels differentiate planned commitments from implemented actions, while separating them from ambiguous, non-committal claims.

\subsection{Annotation Process}
The annotation process involves 5 annotators\footnotemark[\value{footnote}] and 3 verifiers\footnotemark[\value{footnote}], all actively engaged in doctoral or post-doctoral research in sustainability. (1) \textit{Train \& Trial:} Annotators and verifiers undergo rounds of trial annotation, with each round comprising 50 random samples. After each round, annotations are scrutinized for accuracy and conformance to the guidelines, and feedback is provided. The trials occur until each person attains a proficiency of at least 95\% correctly labeled samples. 
(2) \textit{Daily Annotation:} After the trials, annotators label the data daily, and are instructed to flag samples with uncertain annotation. (3) \textit{Resolving Disagreements:} Every 3 days, uncertain annotations are discussed among all annotators to reach a decision. In cases where complete agreement cannot be reached, majority voting is taken. (4) \textit{Validation:} Every 3 days, 20\% of annotated data for each annotator (comprising samples that are not flagged with uncertain annotation) are scrutinized by the verifiers for accuracy and conformance to guidelines. When incorrectly annotated statements exceed 5\%, annotations for the 3-day period are redone.

\begin{table}[h!]
    \centering
    \small
    \renewcommand{\arraystretch}{1.2} % Adjust row height
    \setlength{\tabcolsep}{4.5pt} % Adjust horizontal spacing

    \begin{tabular}{p{6.5cm} r}
        \toprule
        \multicolumn{2}{l}{\textbf{Sustainability Category}} \\
        \midrule
        Resource Optimisation                 & 368  \\
        Emissions Control                      & 284  \\
        Worker \& Consumer Safety Compliance   & 328  \\
        Workplace                               & 401  \\
        Outreach                                & 340  \\
        Management                              & 297  \\
        Business Compliance                     & 167  \\
        Waste Management                        & 307  \\
        Data and Cybersecurity Protection       & 129  \\
        Ecological Conservation                 & 102  \\
        \midrule
        \textbf{Total No. Aspects}             & 666 \\
        \textbf{Total Aspects$^*$}             & 2723 \\
        \midrule
        \multicolumn{2}{l}{\textbf{Action Type}} \\  
        \midrule
        Indeterminate                          & 885  \\
        Implemented                             & 1459 \\
        Planning                                & 379  \\
        \midrule
        \textbf{Total Actions$^*$}             & 2723 \\
        \bottomrule
    \end{tabular}

    \caption{Aspects and action totals for each sustainability category. $^*$ Totals do not include no-aspect counts.}
    \label{tab:aspect_count_and_actions}
\end{table}

\subsection{Dataset Statistics}
% sentence complexities: n-grams, POS, DP, vocab; relative to other famous datasets such as ABSA
A3CG comprises 2004 statements (samples), comparable to standard aspect-level analysis datasets - Rest15~\cite{pontiki-etal-2015-semeval}. Table~\ref{tab:aspect_count_and_actions} summarizes the dataset. Moreover, A3CG follows standard aspect-level analysis datasets - Lapt14, Rest14, 15, 16~\cite{pontiki-etal-2014-semeval,pontiki-etal-2015-semeval,pontiki-etal-2016-semeval}, by having a similar proportion of samples (33.2\%) without aspect-action pairs. This adds the challenge of distinguishing sustainability statements that contain aspect-action pairs from those that do not, simulating real-world statements where relevant aspects are not always mentioned~\cite{khan2016corporatesustainabilitymateriality}.
% Additionally, our dataset aims to have a sizeable number of aspects (>100) for each aspect category~\cite{bai2024bvsp}.

\section{Experiments}

\subsection{Experimental Setups}\label{sec:expt_setup}
\noindent \textbf{Full Dataset\footnote{Appendix~\ref{appendix: expt_details} details all experimental setups and model implementations, including hyperparameter study for AL.}.} As a preliminary step to verify dataset stability before evaluating cross-category generalization, we split the entire dataset into train, validation, and test sets while keeping a balanced ratio of aspect categories. 

\noindent\textbf{Cross-Category Generalization\footnotemark[\value{footnote}].} Different from leave-one-domain-out setups in cross-domain generalization~\cite{xu2023astewild}, cross-category generalization requires a distinct approach. Real-world sustainability statements are complex~\cite{nilssustainabilityreadability2016}, and aspects from different categories tend to co-occur within the same statement, making a leave-one-category-out strategy unrealistic. Therefore, we split the dataset into three equal folds. In each fold, samples are assigned to the training, validation, or test set based on the aspect categories they contain. 3-4 categories are excluded from the training and validation sets, forming the unseen (US) test set. Models are evaluated on the US test set to analyze their capacity for cross-category generalization. A control seen (S) test set is also constructed, comprising categories that overlap with those in the training set, but with entirely different samples. To ensure proper evaluation, no training or validation sample appears in any test set. The set of excluded categories for each fold is chosen considering their tendency to co-occur in a statement and varies across folds with no overlap, enabling balanced evaluation on different unseen categories. 

\subsection{Models}\label{sec:model_grps}

\noindent\textbf{Vanilla Models}\footnotemark[\value{footnote}]: Popular models for aspect analysis are utilized, encompassing generative - T5~\cite{raffel2023t5}, and sequence tagging - BERT Sequence Tagging, BERT-ST~\cite{mao2021absa}.

\noindent\textbf{Vanilla + Learning Paradigm\footnotemark[\value{footnote}]:} 
Contrastive learning (CL) and adversarial learning (AL) are applied to vanilla models - T5+CL, T5+AL, BERT-ST+CL, BERT-ST+AL. CL uses the supervised contrastive loss for pre-training~\cite{li2021contrastivelearning}: $\mathcal{L}_{\mathcal{I}}^{\text{sup}} = 
\sum_{i \in \mathcal{I}} \frac{-1}{|P(i)|} 
\sum_{p \in P(i)} 
\log \frac{\exp(\mathbf{z}_i \cdot \mathbf{z}_p / \tau)}
{\sum_{a \in N(i)} \exp(\mathbf{z}_i \cdot \mathbf{z}_a / \tau)}$.
In batch \(\mathcal{I}\), index \(i\) represents the anchor sample. \(P(i) = \{p \in \mathcal{I} : p \neq i\}\) is the set of indices of all positives distinct from \(i\) (samples with the same category label as the anchor; a sample's category label is determined by the categories of its contained aspect(s)). \(N(i) = \{n \in \mathcal{I} : n \notin P(i)\}\) is the set of indices of all negatives (samples not sharing any category label with the anchor). Following~\citet{ganin2015unsuperviseddomainadvimpl}, AL focuses on learning category-invariant features. A category discriminator is added to the encoder to predict the categories of a sample, with a gradient reversal layer to reverse the discriminator loss gradients.

\noindent\textbf{SOTA-ABSA\footnotemark[\value{footnote}]:} We adapt SOTA-ABSA models for A3CG, given that both ABSA and A3CG tasks overlap in their focus on aspect-level analysis. Therefore, we hypothesize that ABSA methods may be applicable to A3CG, and test this through experiments. We select the latest sequence-tagging methods--GRACE~\cite{Luo2020grace}, and generative methods--InstructABSA~\cite{scaria2023instructabsa}, IT-RER-ABSA~\cite{zheng2024it-rer}, CONTRASTE~\cite{mukherjee2023contraste}.

\noindent\textbf{Large Language Models\footnotemark[\value{footnote}]:} The latest LLMs (zero-shot and few-shot) are evaluated. These are selected from different LLM families, and have achieved state of the art performance on general reasoning and natural language understanding benchmarks. The LLMs evaluated include--GPT-4o~\cite{gpt-4o}, Claude 3.5 Sonnet~\cite{huang2024olympicarenaclaude3.5}, Llama 3 (70B)~\cite{llama3}, DeepSeek V3~\cite{liu2024deepseek}.

\begin{table*}[h]
    \centering
    \small
    \setlength\tabcolsep{6.2pt}
    \begin{tabular}{|c|c|cc|cc|cc|c|c|c|}
        \hline
        \multirow{2}{*}{\textbf{Method}} & \multirow{2}{*}{\shortstack{\textbf{Full} \\ \textbf{Dataset}}} & \multicolumn{2}{c|}{\textbf{Fold 1}} & \multicolumn{2}{c|}{\textbf{Fold 2}} & \multicolumn{2}{c|}{\textbf{Fold 3}} & \multirow{2}{*}{\textbf{S Avg}} & \multirow{2}{*}{\textbf{US Avg}} & \multirow{2}{*}{\textbf{$\Delta$}} \\
        \cline{3-8}
        & & S & US & S & US & S & US & & & \\
        \hline
        T5 & \underline{70.48} & \underline{57.85} & \underline{43.03} & \underline{68.90} & \underline{45.74} & \underline{67.94} & \underline{34.59} & \underline{64.90} & \underline{41.12} & -23.78 \\
        BERT-ST & 43.19 & 39.56 & 25.25 & 39.00 & 22.92 & 43.97 & 30.01 & 40.84 & 26.06 & -14.78 \\
        \hline
        T5 + CL & \underline{71.12} & \underline{\textbf{62.96}} & \underline{46.97} & \underline{69.76} & 46.67 & \underline{67.99} & 38.33 & \underline{66.90} & \underline{43.99} & -22.91 \\
        T5 + AL & 69.27 & 61.24 & 39.62 & 66.91 & \underline{47.02} & 65.17 & \underline{41.82} & 64.44 & 42.82 & -21.62 \\
        BERT-ST + CL & 68.53 & 60.00 & 37.06 & 69.22 & 41.78 & 58.04 & 34.94 & 62.42 & 37.93 & -24.49 \\
        BERT-ST + AL & 24.30 & 37.75 & 27.05 & 27.13 & 25.57 & 34.70 & 26.11 & 33.19 & 26.24 & -6.95 \\
        \hline
        InstructABSA & 69.47 & 60.23 & 37.53 & 64.73 & 49.76 & 64.14 & 47.38 & 63.03 & 44.89 & -18.14 \\
        IT-RER-ABSA & 69.20 & 57.70 & 41.81 & 66.02 & 48.87 & 68.83 & 39.10 & 64.18 & 43.26 & -20.96 \\
        GRACE & 67.09 & 60.87 & \underline{\textbf{50.08}} & 63.10 & 44.33 & 61.92 & \underline{\textbf{48.12}} & 61.96 & \underline{\textbf{47.51}} & -14.45 \\
        CONTRASTE & \underline{\textbf{71.33}} & \underline{61.26} & 48.14 & \underline{\textbf{69.81}} & \underline{\textbf{50.30}} & \underline{\textbf{71.34}} & 40.58 & \underline{\textbf{67.47}} & 46.34 & -21.13 \\
        \hline
        GPT-4o & 29.79 & 31.61 & 42.98 & 40.00 & 32.51 & 35.58 & 32.35 & 35.73 & 35.95 & +0.22 \\
        GPT-4o + FS$^*$ & 35.69 & 39.08 & 46.68 & 39.12 & 41.10 & 40.05 & 33.46 & 39.42 & 40.41 & +0.99 \\
        Claude 3.5 Sonnet & 37.70 & 36.71 & 39.44 & 41.11 & 36.88 & 41.59 & 38.22 & 39.80 & 38.18 & -1.62 \\
        Claude 3.5 Sonnet + FS$^*$ & \underline{42.11} & \underline{40.62} & 46.27 & 43.18 & 40.35 & 45.04 & \underline{39.48} & 42.95 & \underline{42.03} & -0.92 \\
        Llama 3 (70B) & 20.15 & 17.97 & 25.24 & 23.66 & 18.33 & 22.31 & 18.43 & 21.31 & 20.67 & -0.64 \\
        Llama 3 (70B)+ FS$^*$ & 29.82 & 25.03 & 38.11 & 33.49 & 33.30 & 33.15 & 26.65 & 30.56 & 32.69 & +2.13 \\
        DeepSeek V3 & 40.16 & 38.06 & \underline{48.18} & \underline{45.27} & 40.23 & 42.46 & 34.84 & 41.96 & 41.08 & -0.88 \\
        DeepSeek V3 + FS$^*$ & 35.63 & 38.85 & 30.16 & 44.18 & \underline{44.51} & \underline{46.62} & 34.43 & \underline{43.22} & 36.37 & -6.85 \\
        \hline
    \end{tabular}
    \caption{Aspect-Action Analysis (AAA) F1 across full dataset and folds with seen (S) and unseen (US) categories. $^*$FS denotes few-shot, with results averaged over 5 random samplings for the few shot examples. $\Delta$ represents US Avg - S Avg. Best results per model type (following section~\ref{sec:model_grps}) are underlined; overall best results are bolded.}
    \label{tab:full_dataset_and_fold_results}
\end{table*}

\section{Results \& Discussion}
From Table~\ref{tab:full_dataset_and_fold_results}, model evaluation on the full dataset indicates the dataset's stability, with supervised models (T5, T5+CL, CONTRASTE) achieving over 70\% F1. In the following, we focus on A3CG's core objective: aspect-action analysis (AAA) - the extraction of aspect-action pairs in a cross-category generalization setup. Therefore, we primarily discuss performance on unseen (US) categories across all folds. Performance on sub-tasks of AAA is also analyzed: aspect term extraction (ATE) for identifying aspects, and action classification (AC) for classifying action labels after correctly identifying aspects (AC excludes no-aspect cases).

For the models evaluated in our paper, we distinguish between two types of models: supervised models and large language models (LLMs). Supervised models include, from Section~\ref{sec:model_grps}--vanilla models, vanilla + learning paradigm and SOTA-ABSA. These models are based on pre-trained language models (PLMs) (i.e. T5, BERT), which are typically smaller in scale and fine-tuned for specific tasks (in our case, the A3CG task). In contrast, when we refer to LLMs, we refer to larger, general-purpose models (i.e. GPT-4o) that are trained on broad language tasks, and are implemented here via prompt-based methods without any fine-tuning. In the following, we discuss the results of our experiments, by first analyzing the performance of supervised methods (Sections~\ref{sec:supervised_mthd_challenges},~\ref{sec:comparision_lp}), followed by LLMs (Sections~\ref{sec:llm_performance},~\ref{sec:llm_challenges}). At the end, we compare both supervised methods and LLMs (Section~\ref{sec:model_type_comparison}) to contrast the strengths and weaknesses of both methods for A3CG.

\subsection{A3CG Challenges for Supervised Models}~\label{sec:supervised_mthd_challenges}

\begin{figure}[h]
    \centering
    \includegraphics[width=\linewidth]{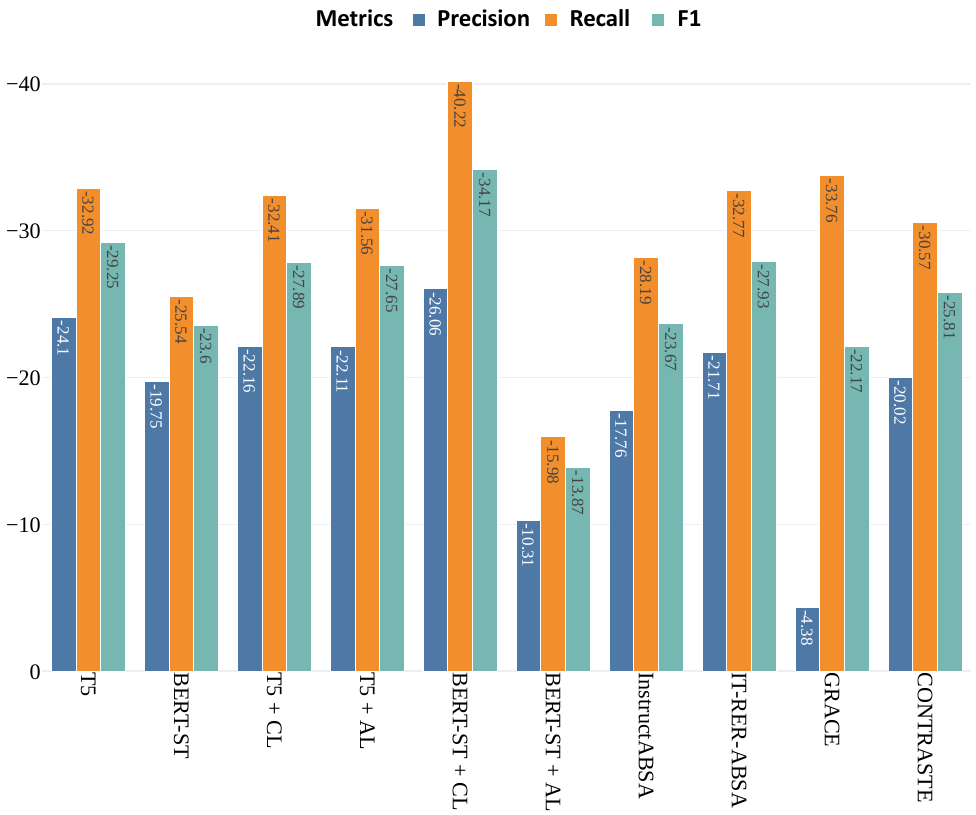}
    \caption{Supervised models' drop in ATE precision, recall, F1, from seen (S) to unseen (US) categories, averaged across all folds (US Avg score - S Avg score).}
    \label{fig:delta_drop}
\end{figure}

\begin{figure*}[t]
    \centering
    \includegraphics[width=\linewidth]{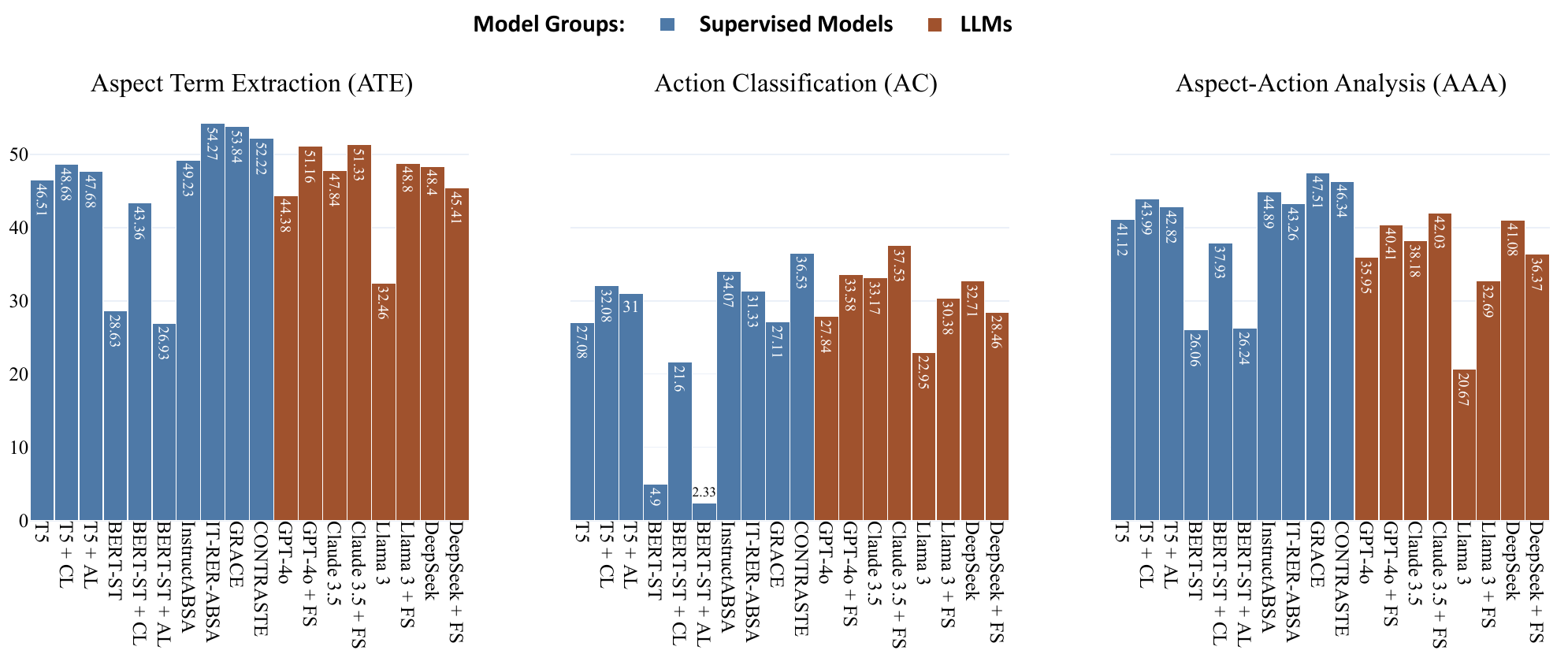}
    \caption{Average F1 score across all folds of unseen categories (US Avg) for supervised and LLM methods. AAA is the main task of A3CG, ATE and AC are subtasks of AAA. For detailed breakdown, refer to Table~\ref{tab:full_AAA_ac_ate_results}.}
    \label{fig:aspect_action_unseen_scores.pdf}
\end{figure*}

\textbf{Supervised models struggle to detect aspects of unseen categories.} From Figure~\ref{fig:delta_drop}, ATE recall drops significantly from seen to unseen categories, averaged across folds. Across all supervised models, the recall decline is notably greater than the precision decline, suggesting that the drop in F1 scores is considerably driven by lower recall. Qualitative analysis indicates that supervised models cannot generalize to sustainability aspects that, while semantically dissimilar to aspects in seen categories, are still highly intuitive to sustainability (i.e. ``carbon emissions''). From Table~\ref{tab:supervised_error_analysis}, these obvious sustainability aspects are typically undetected in ATE. This suggests that models potentially rely more on category-specific patterns instead of a broader conceptual understanding of sustainability-related aspects. Therefore, \textit{to improve generalization, supervised models can be enhanced by incorporating contextually-aware representations~\cite{liang2022senticgcn} that capture broader sustainability concepts.}

% Model performance is also influenced by the vocabulary overlap between the folds, highlighting...

\textbf{Qualitative analysis reveals that supervised models primarily struggle with specific cases of complex syntax.} These prominent failure cases include: i) \textit{Elipsis:} Misclassifying actions when key words that explicate the action taken on the aspect are omitted, with contextual inference required for action classification instead. ii) \textit{Ambiguous Syntax:} When it is unclear how a phrase modifies an aspect, creating ambiguity in the action taken, the model misclassifies the action as a concrete implementation or plan instead of ``indeterminate''. Table~\ref{tab:supervised_error_analysis} shows these examples. Hence, \textit{to improve performance on complex syntax cases, supervised models can incorporate syntactic-aware architectures~\cite{huang2024syntaxaware} to enhance syntactic contextual inference and resolve ambiguous syntax.} 

% i) applies for generative models (models with T5 as the backbone, including T5 implementations within SOTA-ABSA), while ii) and iii) apply for all supervised models. i) \textit{Hallucinations:} Supervised generative methods tend to provide aspects that are not found within the sentence.

% while hallucinations can be mitigated by uncertainty calibration techniques to discard aspects that are semantically misaligned with the input (cite).

\begin{figure*}[h]
    \centering
    \includegraphics[width=\linewidth]{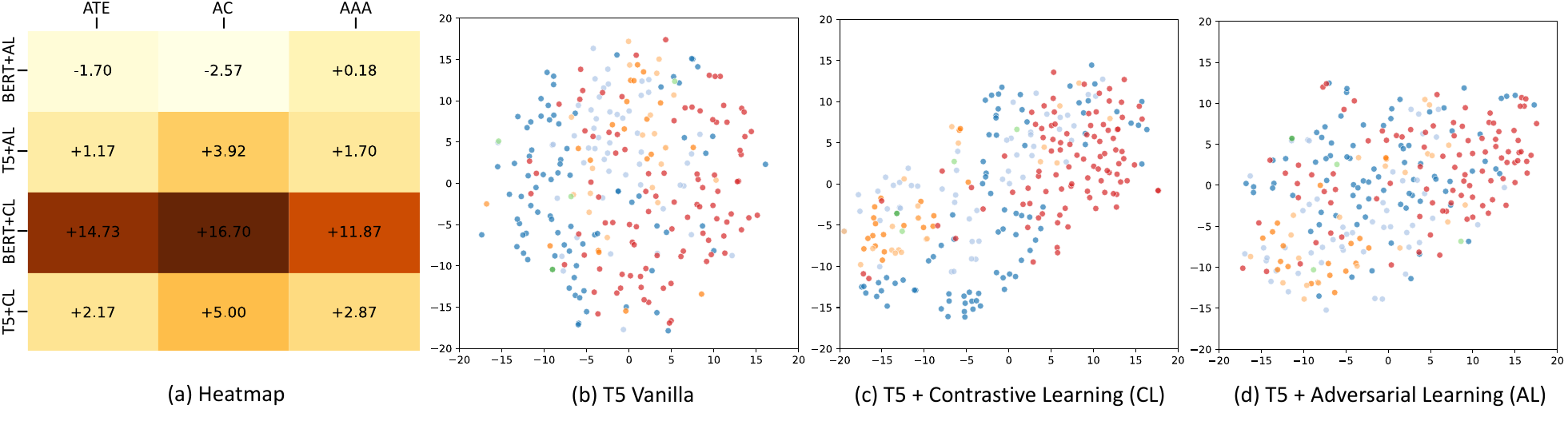}
    \caption{(a) Heatmap of F1 improvements on unseen categories, averaged across all folds, for vanilla+learning paradigm models, compared to original vanilla models. (b), (c), (d) t-SNE visualization of feature embeddings for fold 3 unseen category samples, with samples color-coded by category. Figure~\ref{fig:tsne_embeddings.pdf} shows t-SNE with a full legend.}
    \label{fig:heatmap_tsne_embeddings.pdf}
\end{figure*}

\subsection{Comparison of Learning Paradigms for Supervised Models}~\label{sec:comparision_lp}
Popular methods for cross-domain generalization, adversarial learning (AL) and contrastive learning (CL), are evaluated on cross-category generalization within A3CG. From Figure~\ref{fig:heatmap_tsne_embeddings.pdf}(a), compared to vanilla T5 and BERT-ST baselines, although AL still enhances generalization performance (+1.70 for T5+AL; +0.18 for BERT+AL), CL yields relatively higher F1 improvements on unseen categories (+2.87 for T5+CL; +11.87 for BERT+CL), averaged across all folds.

The generalization effectiveness of CL could be attributed to the models' ability to learn robust semantic features (category distinctions) that are transferable across categories~\cite{khosla2020supervisedcontrastive}. In a case study, the feature embeddings of samples from different unseen categories exhibit more distinct clustering for Figure~\ref{fig:heatmap_tsne_embeddings.pdf}(c) T5+CL, compared to~\ref{fig:heatmap_tsne_embeddings.pdf}(b) T5 Vanilla, which shows greater overlap. This suggests that CL allows the model to capture semantic distinctions between unseen categories, reducing category overlap and ambiguity. Therefore, instead of spurious correlations, the model relies more on robust semantic features unique to each category, improving unseen category generalization~\cite{izmailov2022featurelearncontrastive}. 

In contrast, AL focuses on learning category-invariant features induced by feature collapse to promote generalization~\cite{tang2020advfeatcollapse}. From Figure~\ref{fig:tsne_embeddings.pdf}(d) T5+AL, features from samples of unseen categories appear less discriminable and more entangled, particularly along the y-axis, compared to~\ref{fig:tsne_embeddings.pdf}(b) T5 Vanilla. However, by collapsing features for invariance, AL may inadvertently suppress domain-specific feature attributes that are relevant for identifying aspects of unseen categories. Hence, though AL increases generalization performance, it yields lower improvements than CL.

This presents a key consideration for supervised training techniques in A3CG: \textit{although AL is still beneficial, CL can be a more effective strategy for unseen category generalization.}

\subsection{Comparison of LLM performance}~\label{sec:llm_performance}
\textbf{LLMs with higher reasoning capabilities perform better on A3CG}. From Table~\ref{tab:full_dataset_and_fold_results}, Claude 3.5 Sonnet+FS and DeepSeek V3 yield the highest average F1 on unseen categories among LLMs (US Avg 42.03 \& 41.08 respectively). This aligns with their superior performance over the other LLMs - GPT-4o and Llama 3 (70B) on general reasoning benchmarks~\cite{liu2024deepseek,dubey2024llamaeval}. This suggests that A3CG goes beyond naive text classification to require structured reasoning. Therefore, \textit{selecting LLMs with higher reasoning capabilities can optimize performance on A3CG.}

% \noindent \textbf{Open source models such as DeepSeek V3 perform competitively on A3CG}

\textbf{Few-shot improves performance on nearly all LLMs except DeepSeek V3.} From Figure~\ref{fig:aspect_action_unseen_scores.pdf}, the addition of few-shot examples to LLMs generally improves their F1 performance on AAA for unseen categories. This suggests that LLMs tend to generalize better to unseen categories through in-context learning~\cite{dong-etal-2024-incontextsurvey}. Yet, including few-shot examples for DeepSeek V3 unexpectedly degrades model performance. Preliminary studies on a related model - DeepSeek R1~\cite{guo2025deepseekr1} highlight its over-reliance and failure to generalize from examples~\cite{parmar2025challengesr1fewshot}. Yet, these studies may only have limited applicability to DeepSeek V3, warranting further investigation. Therefore, \textit{to optimize performance on A3CG, few-shot examples can be provided for LLMs such as GPT-4o, Claude 3.5 Sonnet, Llama 3 (70B). In contrast, DeepSeek V3 unexpectedly performs better in zero-shot, underscoring the need for further studies into its few-shot capabilities.}

\begin{figure*}[h]
  \centering
  \begin{subfigure}{0.48\linewidth}
    \includegraphics[width=\linewidth]{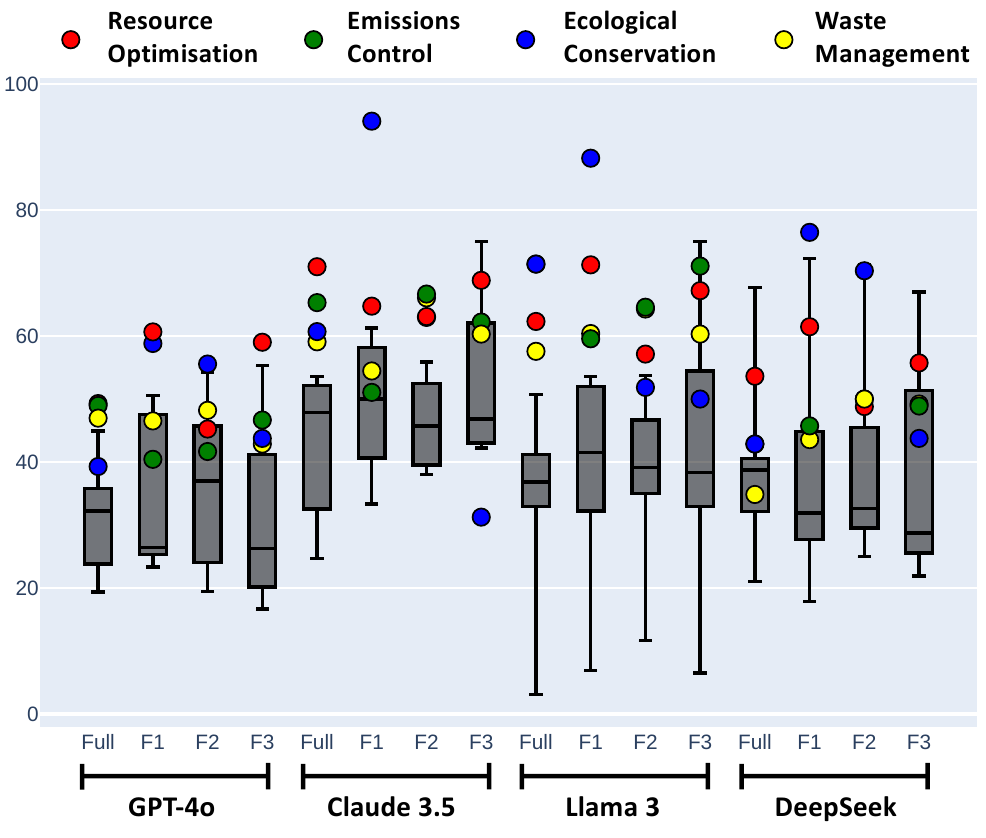}
    \caption{Zero-shot performance}
    \label{fig:llm_zs}
  \end{subfigure} \hfill
  \begin{subfigure}{0.48\linewidth}
    \includegraphics[width=\linewidth]{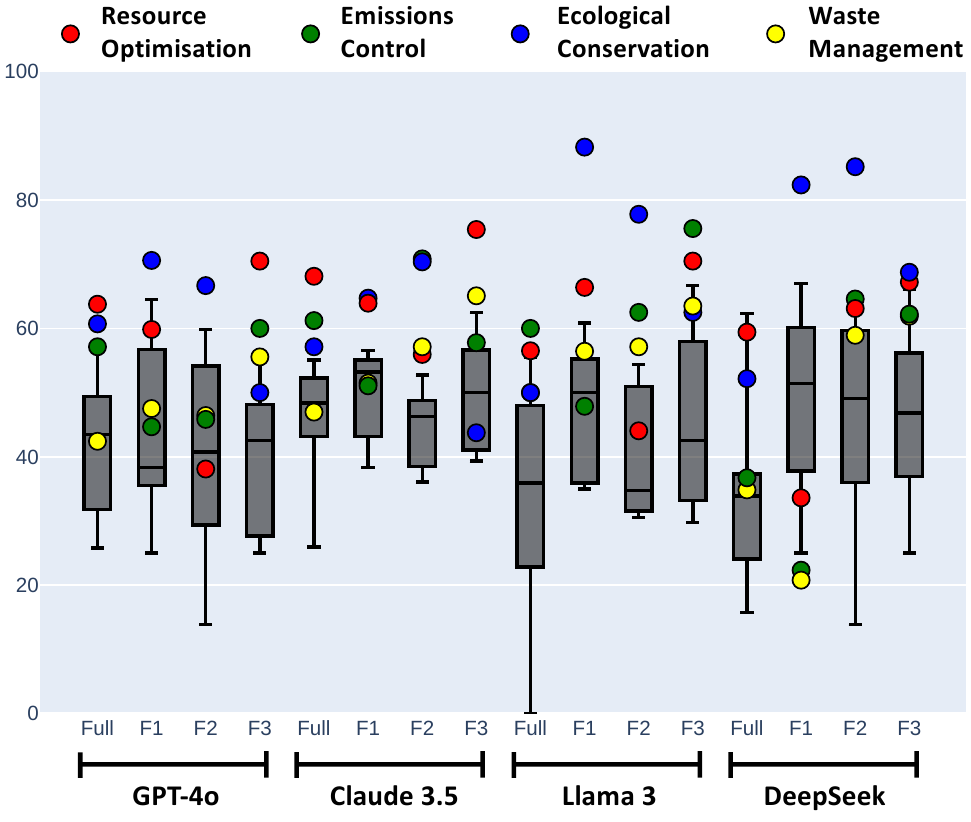}
    \caption{Few-shot performance}
    \label{fig:llm_fs}
  \end{subfigure}
  \caption{ATE recall scores for categories in the Full Dataset (Full) test set and Fold (F1, F2, F3) seen \& unseen test sets. Environmental categories (circles) generally have greater recall than non-environmental categories (box plots).}
  \label{fig:category_recall}
\end{figure*}

\subsection{A3CG Challenges for LLMs}~\label{sec:llm_challenges}
\textbf{LLMs generally exhibit higher recall toward aspects of environmental categories, potentially due to pre-training biases.} Figure~\ref{fig:category_recall} shows that ATE recall on all environmental categories (\textit{resource optimization, emissions control, ecological conservation}) tend to exceed the median ATE recall of non-environmental categories, frequently surpassing their upper quartile. This suggests that LLMs capture environmental aspects more effectively than non-environmental ones, potentially due to popularity biases in their pre-training~\cite{dai2024llmbiassurvey}. Since LLMs are pre-trained on large-scale user-generated data, their learned distributions are often influenced by the popularity of specific topics - in this case, the prominent association of sustainability with environmental protection over non-environmental areas (i.e. social and governance)~\cite{ruggerio2021sustainabilityprominentenv}. Consequently, LLMs may end up disproportionately favoring environmental aspects for ATE while under-representing non-environmental sustainability aspects. However, while our findings highlight this possibility, further research is needed to assess how much these biases contribute to the recall skew. Therefore, \textit{to mitigate this imbalance for improved performance, methods such as bias-aware re-ranking~\cite{carraro2024enhancingllmdiv} can be employed to balance the selection of non-environmental aspects, although deeper investigation into the causes of the recall skew could enhance mitigation strategies.}

\textbf{Qualitative analysis reveals that LLMs primarily struggle with specific cases of complex pragmatics}. Pragmatics deals with the implicit elements of language that convey intent and contextual nuance beyond literal meaning~\cite{MAO2025pragmatic}. In both zero-shot and few-shot settings, LLMs' most prominent failure cases involve pragmatic challenges that affect the intent, commitment, and certainty of sustainability claims: i) \textit{Misinterpreting modality:} Misclassifying actions as concrete commitments or definite plans, despite the statement containing tentative and hedging language that indicate uncertainty. ii) \textit{Negation handling:} Misclassifying negative actions as implementations or plans to improve sustainability. iii) \textit{Unattributed statements:} Misclassifying actions as concrete company-specific implementations or plans, despite the statement lacking explicit attribution to the company. iv) \textit{Future Dependency:} Misclassifying planned actions as already implemented, despite the statement emphasizing a future dependency required for the action to occur. Table~\ref{tab:llm_error_analysis} details these examples. Therefore, \textit{to adapt LLMs for capturing pragmatic cues, prompting strategies such as chain-of-thought~\cite{wei2022cot} and self-reflection~\cite{ji2023selfreflection}, can be employed to explicitly reason through \textit{modality, negation, attribution, future dependency} before classification, thereby improving robustness on these pragmatic cases.}

\subsection{Comparison of Model Types}\label{sec:model_type_comparison}
\textbf{Supervised models can perform better than LLMs on unseen categories}, with GRACE achieving the highest average F1 score on unseen categories (US Avg 47.51) compared to LLMs, from Table~\ref{tab:full_dataset_and_fold_results}. As shown in Figure~\ref{fig:aspect_action_unseen_scores.pdf}, supervised models generally outperform LLMs in AAA on unseen categories, except for T5, BERT-ST, BERT-ST+CL, BERT-ST+AL. This suggests that while LLMs perform well on a broad range of tasks, supervised fine-tuning is generally better at capturing task-specific patterns within A3CG. Thus,~\textit{supervised fine-tuning remains essential for optimal performance on A3CG.}

\textbf{While supervised models have higher absolute performance than LLMs, they show significant inefficiency at transferring performance from seen to unseen categories.} For supervised models, although they demonstrate the best absolute performance on unseen categories, the difference between average F1 scores on seen and unseen categories, $\Delta$, is significant - greater than 14\% in magnitude except BERT-ST+AL ($\Delta$=-6.95\%), from Table~\ref{tab:full_dataset_and_fold_results}. This represents a substantial drop in performance when transferring learned patterns from seen to unseen categories, underscoring the weak generalization efficiency of learned features. The features learned by LLMs are shaped by broad pre-training, making them generally applicable across diverse data distributions. As a result, the performance drop on unseen categories is relatively modest, as these categories do not represent truly out-of-distribution data. In contrast, supervised models rely on features tuned to the training distribution, and their performance shortfall on unseen categories reflects a stronger sensitivity to distribution shifts. Consequently, \textit{supervised models can improve the generalization efficiency of their learned features, to enhance cross-category generalization.}

\textbf{Specific LLMs can still perform competitively on unseen categories.} Despite being generally outperformed by supervised models, specific LLMs achieve competitive average F1 scores on unseen categories, narrowing the performance gap. From Figure~\ref{fig:aspect_action_unseen_scores.pdf}, Claude 3.5 Sonnet+FS (42.03) outperforms several supervised models - T5 (41.12), BERT-ST (26.06), BERT-ST+CL (37.93), BERT-ST+AL (26.24), while lagging behind IT-RER-ABSA (43.26), T5+AL (42.82) by less than 1.5\%. Therefore, \textit{while supervised learning provides optimal performance, its superiority is not absolute - specific LLMs can still present viable alternatives to full-scale supervised fine-tuning, particularly in scenarios where computational resources are limited but API-based access to LLMs are available.}

\textbf{Action classification is more challenging than aspect term extraction for all methods.} From Figure~\ref{fig:aspect_action_unseen_scores.pdf}, AC F1 scores are markedly lower than ATE on unseen categories across all methods. This is potentially because while both subtasks involve semantic, syntactic and pragmatic reasoning, AC places greater emphasis on interpreting complex syntax and pragmatics, as discussed in the failure cases in supervised models (Section~\ref{sec:supervised_mthd_challenges}) and LLMs (Section~\ref{sec:llm_challenges}). Hence, AC is particularly challenging for supervised and LLM approaches. Therefore, \textit{given the considerable difficulty and lower performance, targeted model improvements for AC could drive significant gains in the overall performance of AAA.} 

\textbf{Comparative error analysis reveals that LLMs excel in syntax but struggle with pragmatics, while supervised models display the reverse tendency}. Unlike supervised models, LLMs are largely unaffected by syntactic challenges (\textit{elipsis}, \textit{ambiguous syntax}) in A3CG, possibly due to their broad syntactic exposure from large-scale pre-training~\cite{yang2024llmsyntaxpretrain}. However, relative to supervised models, LLMs struggle with specific pragmatic cases (\textit{modality, negation, unattributed statements, future dependency}) that shape how sustainability claims are framed. This suggests that relying solely on pre-training constrains LLMs' pragmatic reasoning capabilities, limiting their ability to interpret the nuanced, context-dependent discourse in A3CG. Without fine-tuning, LLMs tend to rely on dominant linguistic patterns from pre-training, rather than aligning with the pragmatic cues of a given context~\cite{ruis2024goldilocks}. This underscores a key point: \textit{while prompting techniques can elicit reasoning in pragmatic cases, they remain susceptible to patterns in the pre-training distribution. Supervised fine-tuning, in contrast, adjusts the training distribution to more effectively internalize pragmatic cues, allowing more reliable contextual interpretation. This is significant in A3CG, let alone the greenwashing context, where pragmatic cases like hedging, indirect intent, strategic ambiguity, are prevalent~\cite{siano2017greenwashingmarkers}.}

 % i) \textit{Greedily extracting aspects:} Incorrectly extracting aspects irrelevant to sustainability.

% \begin{table}[b]
%     \centering
%     \small
%     \setlength\tabcolsep{2pt}
%     \begin{tabular}{|c|c|c|c|c|}
%         \hline
%         \textbf{Method} & \textbf{Fold 1} & \textbf{Fold 2} & \textbf{Fold 3} & \textbf{Avg} \\
%         \hline
%         GPT-4o + FS(S) & 42.29 & 33.42 & 28.57 & 34.76 \\
%         Claude 3.5 Sonnet + FS(S) & 41.18 & 38.95 & 39.68 & 39.94  \\
%         Llama 3 (70B) + FS(S) & 30.15 & 25.17 & 33.06 & 29.46 \\
%         DeepSeek V3 + FS(S) & 30.33 & 38.77 & 33.17 & 34.09 \\
%         \hline
%     \end{tabular}
%     \caption{F1 results for Source FS-enabled methods with unseen (US) data across folds and their averages. Results are taken from the best of 5 random samplings for few shot.}
%     \label{tab:source_few_shot}
% \end{table}

\section{Conclusion}
In this study, we discussed the greenwashing phenomenon, and highlighted significant limitations of current NLP methods for robust ESG analysis. To address this, we proposed the A3CG dataset and evaluated supervised models, and LLMs under zero and few-shot settings. While supervised models achieved the highest absolute performance for unseen category generalization, LLMs exhibited greater generalization efficiency from seen to unseen categories. Within supervised approaches, contrastive learning proved more effective than adversarial learning for unseen category generalization. Yet both supervised and LLM approaches face limitations. On the one hand, supervised models lack contextual awareness for detecting aspects of unseen categories and struggle to classify actions in complex syntactic cases. On the other hand, LLMs overlook non-environmental aspects possibly due to pretraining biases, and struggle with classifying actions where there is pragmatic nuance. These results underscore the need for further research into models that can generalize robustly across ESG categories for the aspect-action analysis task. Therefore, we invite researchers to experiment with A3CG to unlock the potential of NLP for robust ESG analysis.

\section*{Limitations}
The A3CG dataset focuses on sustainability statements that are solely in English. However, we acknowledge that sustainability is a global responsibility that can encompass many different regions. Therefore, for greater inclusivity, future work will extend A3CG to non-English languages.  Additionally, due to computational resources, this paper does not consider supervised fine-tuning of smaller LLMs--i.e. Llama 3 8B~\cite{llama3}. Yet, LoRA-based fine-tuning of these smaller scale LLMs has proven effective in other domain-specific tasks~\cite{varshney2025climaempact}, and could represent a future research direction.

\section*{Ethical Considerations}
Procedures for data collection were approved by an internal ethics review board within our research group. Additionally, we adhere to ethical principles by ensuring that all data collection and processing are performed with respect for privacy and confidentiality. We use publicly available sustainability disclosures from respective company websites. Considering that these disclosures may contain company and personal names, we make efforts to anonymize sensitive and personal information in the A3CG dataset, focusing solely on the sustainability content relevant to our research. Additionally, the models utilized are publicly available, found from published research papers. Our usage of all data, packages and models adheres to the copyright guidelines provided by the respective copyright holders. Human annotators follow strict guidelines to maintain objectivity and reduce bias. We also ensure transparency in our methodology and provide clear attribution for our sources, aiming to support ethical practices in data usage and dissemination. The dataset and experimental code used in this study will be made publicly available upon acceptance, for the sole purpose of facilitating research, accompanied by appropriate copyright provisions.

\section*{Acknowledgements}
This research/project is supported by the NUS Sustainable and Green Finance Institute (SGFIN), NUS Asian Institute of Digital Finance (AIDF), Ministry of Education, Singapore under its MOE Academic Research Fund Tier 2 (STEM RIE2025 Award MOE-T2EP20123-0005: “Neurosymbolic AI for Commonsense-based Question Answering in Multiple Domains”), MOE Tier 2 Award (MOE-T2EP50221-0006: ‘‘Prediction-to-Mitigation with Digital Twins of the Earth’s Weather’’), MOE Tier 1 Award (MOE-T2EP50221-0028: ‘‘Discipline-Informed Neural Networks for Interpretable Time-Series Discovery’’), and by the RIE2025 Industry Alignment Fund – Industry Collaboration Projects (IAF-ICP) (Award I2301E0026: “Generative AI"), administered by A*STAR, as well as supported by Alibaba Group and NTU Singapore.

\bibliography{main}

% \clearpage
\appendix

\section{Dataset}\label{sec:humanannotate}
\subsection{Dataset Samples}
Additional samples from the A3CG dataset are presented in Table~\ref{tab:dataset_examples}, covering sustainability categories not shown in Figure~\ref{fig:overview_a3cg}. This provides an appreciation of samples from all sustainability categories, including samples that comprise aspects from multiple categories.

\subsection{Annotators}\label{sec:humanannotators}
There are a total of 5 human annotators and 3 human verifiers, all based in Singapore and recruited from reputable research institutions. These annotators and verifiers are affiliated with the Asian Institute of Digital Finance and the National University of Singapore College of Design and Engineering. The annotators and verifiers are actively engaged in doctoral or post-doctoral research in the corporate sustainability field, with expertise in sustainability analysis. They contribute to the annotation process as part of their formal academic and research activities, and are compensated above the local minimum wage (SGD\$15/hr) for their involvement. They strictly follow the annotation scheme and guidelines while annotating the dataset, and have consented to its use for research purposes.

\subsection{Annotation Instructions}
This section details the full text of instructions given to all annotators. Each annotator is instructed to read the background, before the general instructions, aspect and action annotation guidelines.\\

\noindent\textbf{Background:}
Sustainability reports are important for evaluating a company's environmental, social and governance (ESG) performance. These reports are regularly released by companies, and contain sustainability statements that describe the company's sustainability efforts. For this annotation task, you will be analyzing sustainability statements to annotate them for research purposes. \\

\noindent\textbf{General Instructions:}

\begin{enumerate}
    \item Please annotate (aspect, action) pairs within sustainability statements.
    \item For each pair, the aspect and actions are connected. This means that the aspect provides a focal point for the action to address or engage with.
    \item Annotate the aspect and actions according to the respective aspect and action guidelines that we will provide.
    \item Each aspect can be classified under a specific sustainability category that it belongs to. Please output the associated category for each aspect. 
    \item If there are no aspects within the sustainability statement, annotate the statement with a (''no aspect'', ``no action'') label. 
    \item For sustainability statements where you are uncertain about the annotation, please flag them up for discussion with the rest of the reviewing team. 
\end{enumerate}

\noindent\textbf{Guidelines for Aspect Annotation:}
\begin{enumerate}
    \item An aspect describes a part or attribute of a specific sustainability category, that is explicitly found within the sustainability statement.
    \item In our case, an aspect can either be an~\textit{entity, goal, sub-area, activity} of a selected sustainability category, providing a focal point for an action to address or engage with. Examples include, ``solar panels'' (entity of \textit{resource optimization}), ``carbon reduction targets'' (goal of \textit{emissions control}), ``wildlife conservation'' (sub-area of \textit{Ecological Conservation}), ``workplace safety sessions'' (activity of \textit{Worker \& Consumer Safety}).      
    \item We provide a list of sustainability categories that an aspect can belong to. These include the categories of \textit{Resource Optimization, Waste Management, Emissions Control, Ecological Conservation, Workplace, Outreach, Management, Business Compliance, Worker \& Consumer Safety, Data \& Cybersecurity Protection}. [Note: We provided the annotators with Table~\ref{tab:taxonomy}, which provides a full explanation of all the sustainability categories.]
    \item The aspect must directly and explicitly relate to a specific sustainability category, requiring no additional assumptions to establish its relevance.
    \item Where an aspect comprises multiple words, it must be a continuous sequence of words within the sustainability statement. Therefore, a single aspect cannot appear in two or more different parts of the sentence. 
    \item As much as possible, an aspect should be described with greater specificity, including additional terms such as qualifiers. An example of including qualifiers is the annotation of ``workplace safety measures'' instead of ``safety measures''.
    \item However, to maintain brevity, an aspect should avoid redundant words that do not add to its core meaning. For example, ``safety protocols'' should be annotated instead of ``safety protocols in the event of similar incidents''.

\end{enumerate}

\noindent\textbf{Guidelines for Action Annotation:}

\begin{enumerate}
    \item Action labels characterize the type of action that is taken toward the aspect based on the sustainability statement. For each aspect, these action labels constitute one of the following: \textit{planning, implemented, indeterminate}. 
    \item \textbf{Planning}: Indicates that an action has been planned or a commitment has been made by a company to address or engage with the aspect to advance its sustainability efforts. This can include:
    \begin{itemize}[leftmargin=*]
        \item Future plans to incorporate a sustainability aspect within sustainability operations, particularly in cases where the aspect is an activity or entity. For example, in the statement, ``the company plans to increase waste recycling by 50\% in 2025'', ``waste recycling'' is the aspect, and ``planning'' is its associated action.
        \item Future commitment with respect to a sustainability aspect, particularly in cases where the aspect is a goal or sub-area. For example, in the statement, ``the group is firmly committed to improve the local community welfare this year'', ``local community welfare'' is the aspect, and ``planning'' is its associated action.
    \end{itemize}
    \item \textbf{Implemented}:
    Indicates that an action has already been taken to address or engage with the aspect to advance its sustainability efforts. This can include:
    \begin{itemize}[leftmargin=*]
        \item A sustainability aspect has already been incorporated within sustainability operations, particularly in cases where the aspect is an activity or entity. For example, in the statement, ``the company has successfully deployed wastewater filtration technologies'', ``wastewater filtration technologies'' is the aspect, and ``implemented'' is its associated action.
        \item A sustainability aspect has been advanced or achieved, particularly in cases where the aspect is a goal or sub-area. For example, in the statement, ``we have successfully achieved 20\% lower carbon emissions in 2024'', ``carbon emissions'' is the aspect, and ``implemented'' is its associated action.
    \end{itemize}
    \item \textbf{Indeterminate}: Indicates that it is unclear from the statement if the company intends to address or engage with the aspect to advance its sustainability efforts, or how it intends to do so. This can include:
    \begin{itemize}[leftmargin=*]
        \item Statements without clear attribution as to who is engaging with the aspect, casting doubt over whether the company is explicitly involved in the engagement. For example, in the statement ``Investments are being made in the country's renewable energy projects'', ``renewable energy projects'' is the aspect, while ``indeterminate'' is its associated action.
        \item Statements that reflect uncertain or non-committal language with respect to engaging with the aspect, lacking the clarity required to indicate concrete commitments, plans or implementations. These include statements that possess hedging language that express uncertainty, without assurance of follow-through. For example, the hedging phrase, ``if feasible'', in the statement, ``we intend to improve workplace diversity, if feasible''. For this statement, ``workplace diversity'' is the aspect, while ``indeterminate'' is its associated action. Other examples include the use of exploratory language, again, without any assurance of follow-through, such as ``we are considering'' in the statement ``we are considering increasing the funding for our community development program''. For this statement, ``community development program'' is the aspect, while ``indeterminate'' is its associated action. 
        \item Statements that do not highlight an enhancement in sustainability. These can include mere disclosures of figures, such as, ``we will disclose greenhouse emission numbers from next year'', where ``greenhouse emission numbers'' is the aspect, and ``indeterminate'' is its associated action. Additionally, simply listing responsibilities without an explicit link to improvements in sustainability. For example, ``the board is responsible for corporate governance'', where ``corporate governance'' is the aspect, and ``indeterminate'' is its associated action.
    \end{itemize}

\end{enumerate}

\section{Sustainability Categories}\label{appendix: sustainability_categories}
Table~\ref{tab:taxonomy} shows the sustainability categories of the A3CG dataset, as well as their corresponding definitions. These categories align with standard sustainability taxonomies such as SASB~\cite{Eng2022sasb} to remain relevant to corporate sustainability stakeholders. 

\section{Experiments}\label{appendix: expt_details}
\subsection{Full Details on Data Partitions}
Table~\ref{tab:data_split_num_samples} highlights the categories that constitute the different data splits discussed in section~\ref{sec:expt_setup}, and specifies the number of samples for each data partition. For each fold, seen categories correspond to the categories found in the seen train, seen val, seen test sets, while unseen categories correspond to the categories found in the unseen test set. Notably, the seen and unseen categories vary across folds to ensure a balanced evaluation. Additionally, the samples are split to ensure a train-validation-test ratio of roughly 68:11:21, with the test set comprising either seen or unseen categories. 

We further breakdown the different aspect counts corresponding to the categories in Table~\ref{tab:data_split_num_aspects}. For each partition of train, validation, test set, we ensure that no aspect samples constitute roughly 22\% of the total count of aspect + no aspects. Consistent with popular aspect-based analysis SemEval datasets~\cite{pontiki-etal-2014-semeval, pontiki-etal-2015-semeval, pontiki-etal-2016-semeval}, the inclusion of no aspect samples simulates real-world conditions, whereby some sustainability statements may not contain any aspects~\cite{khan2016corporatesustainabilitymateriality}. Therefore, models encounter the realistic challenge of distinguishing sustainability statements that contain aspect-action pairs from those that do not, improving the robustness of our study.

Models are trained and validated exclusively on seen categories (i.e. the seen train and seen validation sets) only, and are then tested on unseen test sets. Further evaluation on the seen test set serves as a control experiment. This setup effectively prevents data leakage from the unseen categories during testing, ensuring the robustness of the results for unseen category generalization.

% Accordingly, we split the dataset into three equal folds. In each fold, samples are split based on the categories associated with the aspects they contain. To create the unseen (US) test set, 3-4 categories are excluded from the training and validation sets. Models are then tested on the US test set to analyse their ability for cross category generalization. We also construct a control seen (S) test set, which includes samples from categories present during training. Importantly, no training samples appear in any test sets. The excluded categories are chosen based on their tendency to co-occur in a statement and vary across folds with no overlap, enabling evaluation on different unseen categories. Folds are used instead of the popular leave-one-out setup (cite) to better reflect the real-world co-occurrence of aspect from different categories within a statement.

\begin{figure*}[h]
    \centering
    \includegraphics[width=\linewidth]{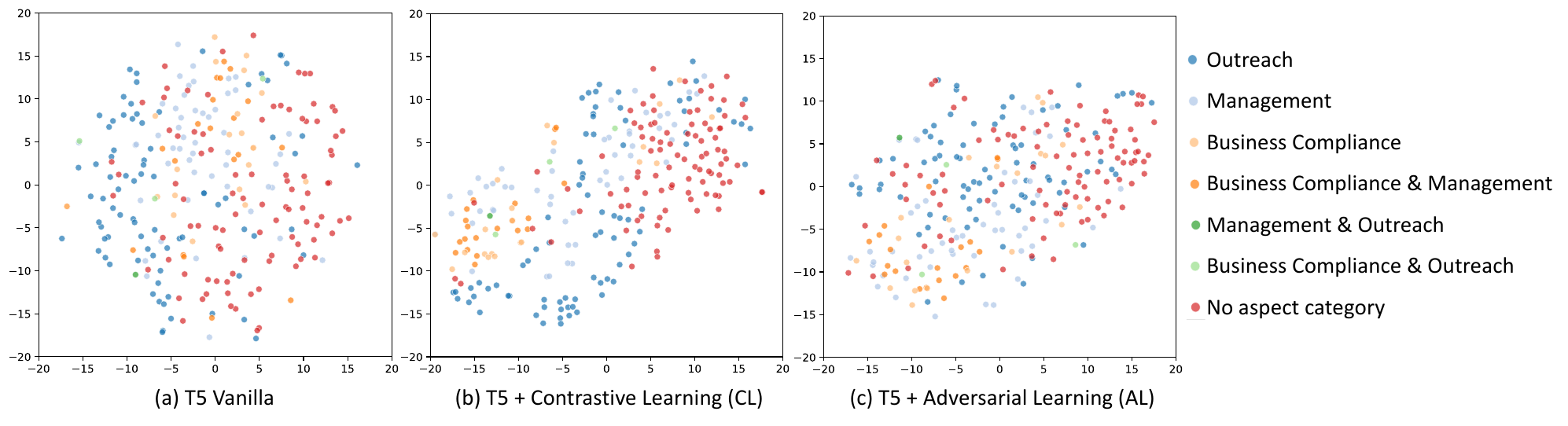}
    \caption{(Version with legend) t-SNE visualization of feature embeddings for fold 3 unseen samples, with each sample color-coded by category.}
    \label{fig:tsne_embeddings.pdf}
\end{figure*}

% \begin{figure}[h]
%     \centering
% \includegraphics[width=0.8\linewidth]{figures/heatmap_deltas.pdf}
%     \caption{Unseen Avg. F1 Improvements for Contrastive Learning (CL) and Adversarial Learning (AL) over T5 \& BERT.}
%     \label{fig:heatmap_improvement.pdf}
% \end{figure}

\subsection{Model Implementation Details} 
To adapt the SOTA-ABSA methods - InstructABSA~\cite{scaria2023instructabsa}, IT-RER-ABSA~\cite{zheng2024it-rer}, GRACE~\cite{Luo2020grace}, CONTRASTE~\cite{mukherjee2023contraste}, the prediction target is changed to that of the aspect-action (aspect, action) pair in place of (aspect, sentiment). For CONTRASTE, the opinion terms are omitted in the prediction target, as well as the Opinion Term Detection module. For InstructABSA and IT-RER-ABSA, we leverage the original format of the prompt for the Sentiment-Pair Extraction (ASPE), adapting the prompt for aspect-action classification. All the prompts utilized in supervised models (InstructABSA, IT-RER-ABSA, T5, T5+AL, T5+CL,) can be found in Table~\ref{tab:method_prompts}, while the corresponding few-shot examples for InstructABSA can be found in Table~\ref{tab:few_shot_examples_instructabsa}. Few-shot examples for IT-RER-ABSA are retrieved from the training sets through the retrieval module in its original implementation. For vanilla models (T5, BERT-ST), as well as their implementation with learning paradigms (T5+AL, T5+CL, BERT-ST+AL, BERT-ST+CL), we use the T5-base\footnote{https://huggingface.co/google-t5/t5-base} and BERT-base-uncased\footnote{https://huggingface.co/google-bert/bert-base-uncased} pre-trained models from HuggingFace\footnote{https://huggingface.co}. The implementation of our sequence tagging framework in BERT-ST, BERT-ST+CL, BERT-ST+AL, follows the joint BIO scheme in~\cite{mao2021absa}, and is summarized in Table~\ref{tab:seq_tag_bert_st}, which shows the extraction of the aspect-action pair (carbon emissions, implemented), from the input ``our carbon emissions have reduced.''

\begin{table}[h]
    \centering
    \small
    \begin{tabular}{cccccc}
        \toprule
        Input & our & carbon & emissions & have & reduced \\
        \midrule
        Aspect & O & B & I & O & O \\
        Action & O & IMP & IMP & O & O \\
        \bottomrule
    \end{tabular}
    \caption{Sequence Tagging using joint-BIO scheme. This example highlights the prediction of the aspect-action pair (carbon emissions, implemented). }
    \label{tab:seq_tag_bert_st}
\end{table}

For all LLMs utilized - GPT-4o~\cite{gpt-4o}, Claude 3.5 Sonnet~\cite{huang2024olympicarenaclaude3.5}, Llama 3 (70B)~\cite{llama3}, and DeepSeek V3~\cite{liu2024deepseek}, we leverage the prompt template in Table~\ref{tab:method_prompts}. We provide few-shot examples in Table~\ref{tab:few_shot_examples_llm} for the LLM few-shot setups. Few-shot examples are selected from the same categories as the test set but are not drawn from the test set itself. For consistency, the few-shot examples are kept the same across all LLMs tested. The versions of LLMs utilized include: gpt-4o-2024-08-06 for GPT-4o, claude-3-5-sonnet-20241022 for Claude 3.5 Sonnet, llama3-70b-8192 for Llama 3 (70B), DeepSeek-V3 2024/12/26 for DeepSeek V3.

\subsection{Hyperparameters for Supervised Models}
To ensure the fairness and robustness of results~\cite{xu2023astewild}, we adopt the hyperparameter settings specified in the original papers for the ABSA-SOTA models - GRACE~\cite{Luo2020grace}, IT-RER-ABSA~\cite{zheng2024it-rer}, InstructABSA~\cite{scaria2023instructabsa}, CONTRASTE~\cite{mukherjee2023contraste}. This ensures that the models reflect their intended configurations and design choices. For T5 and BERT-ST, as well as their respective learning paradigm implementations (T5+AL, T5+CL, BERT-ST+AL, BERT-ST+CL), a learning rate of 3e$^{-5}$ was utilized, with a batch size of 8. For T5+CL and BERT-ST+CL, the models were pre-trained using contrastive learning, with a temperature, $\tau$, of 0.07, a maximum number of positives and negatives of 2 each, and pre-training over 20 epochs with early stopping (patience of 3). For adversarial training setups (T5+AL, BERT-ST+AL), the model might be highly sensitive to the parameter $\alpha$, which denotes the gradient scaling factor used in the gradient reversal layer (i.e. the strength of adversarial signal). For adversarial setups, we utilize a reasonable default $\alpha$ of 0.3, while conducting a hyperparameter study on $\alpha$ in appendix~\ref{appendix:advablation}. All models are trained over 100 epochs on the training set, where the model instance with the best score on the validation set is used for test set evaluation. 

\subsection{Ablation Study for Adversarial Setups}\label{appendix:advablation}
\begin{equation}
\nabla_{\theta} \mathcal{L} = \nabla_{\theta} \mathcal{L}_{\text{task}} - \alpha \nabla_{\theta} \mathcal{L}_{\text{adv}}
\end{equation}

% \begin{equation}
% \mathcal{L} = \mathcal{L}_{\text{task}} + \mathcal{L}_{\text{adv}}
% \end{equation}

In this study, the implementation of adversarial learning methods (T5+AL, BERT-ST+AL) follows from~\cite{ganin2015unsuperviseddomainadvimpl}, where the objective is for the encoder to learn category-invariant features. First, a category discriminator is added to the encoder to predict the categories present in a given sample. Then, a gradient reversal layer is added to reverse the gradients resultant from the discriminator loss. We summarize this in equation (1) where $L_{\text{adv}}$ represents the discriminator loss, and $\alpha$ represents the gradient scaling factor which controls the strength of the adversarial signal. The total loss $L$ combines the task loss $L_{\text{task}}$ with a scaled $L_{\text{adv}}$, to ensure that the encoder learns category-invariant representations while balancing the learning of the primary task. Following from~\cite{ganin2015unsuperviseddomainadvimpl}, $\alpha$ controls the trade-off between the primary task objective and the adversarial objective (learning category-invariant features). 

Since the main results of this study focus on the performance on unseen categories, tuning $\alpha$ on the validation set (comprising seen categories) would not necessarily optimize performance on the unseen test set (comprising unseen categories). Therefore, we selected a default $\alpha$ value of 0.3 for all adversarial setups (T5+AL, BERT-ST+AL). To validate the robustness of this choice, we observe how varying $\alpha$ impacts F1 for BERT-ST and T5 on the unseen test set. We examine
$\alpha$ $\in$ \{0.01, 0.05, 0.1, 0.2, 0.3, 0.4, 0.5, 0.6, 0.7, 0.8, 0.9, 1.0, 1.1, 1.2, 1.3, 1.4, 1.5, 1.6, 1.7, 1.8, 1.9, 2.0\}, a range that broadly aligns with foundational adversarial learning setups~\cite{ganin2016domainadvlearning}. 

\begin{figure}[h]
    \centering
    \includegraphics[width=\linewidth]{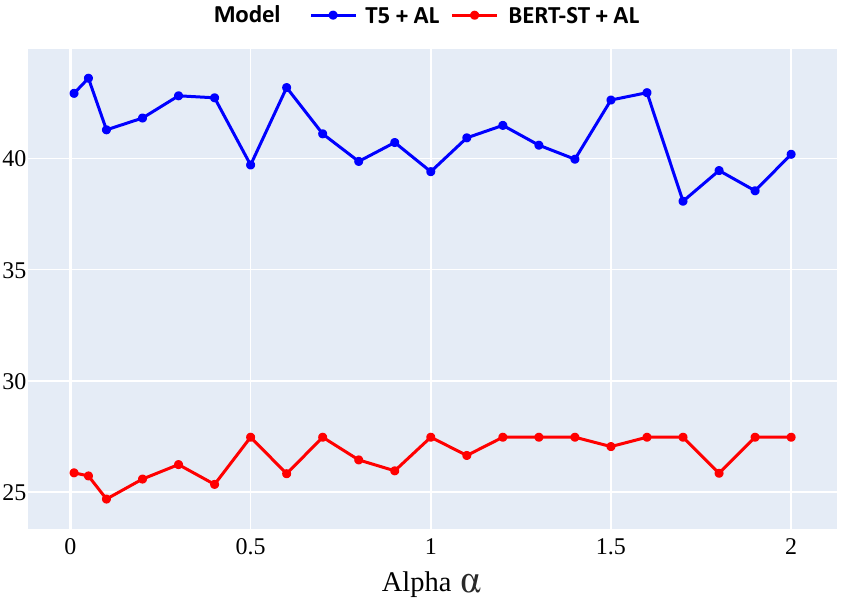}
    \caption{F1 results averaged across unseen test sets for all folds (US Avg),  for T5+AL and BERT-ST+AL, corresponding to different values of $\alpha$.}
    \label{fig:ablation_alpha_test_f1}
\end{figure}

From Figure~\ref{fig:ablation_alpha_test_f1}, we observe that even in an ideal scenario where the chosen $\alpha$ value maximizes performance across all unseen test sets, the US Avg (average f1 across unseen (US) test sets of all folds) only reaches a maximum value of 43.19 for T5+AL ($\alpha$ = 0.6) and 27.47 for BERT-ST+AL ($\alpha$ = 0.5). However, note that this represents only the ideal $\alpha$ value, which may not be achievable in practice without tuning on the unseen test sets - an approach that would constitute an unfair implementation. Moreover, even with these ideal $\alpha$ values, the observed gains over the experimentally selected $\alpha$ value of 0.3 - which yielded 42.82 for T5+AL and 26.24 for BERT-ST+AL (Table~\ref{tab:full_dataset_and_fold_results}), remain marginal and are still lower than the results achieved with T5+CL and BERT-ST+CL. As a result, the ablation study underscores that while adversarial learning still provides benefits for cross-category generalization in A3CG, contrastive learning is relatively more effective. This aligns with our discussion in section~\ref{sec:comparision_lp}. We provide the full breakdown of US Avg F1 results for different $\alpha$ in Table~\ref{tab:alpha_performance}.

\subsection{Error Analysis Examples}
All error examples discussed in the main text are outlined in tables~\ref{tab:supervised_error_analysis} and~\ref{tab:llm_error_analysis}

\subsection{Results Breakdown}
The detailed results breakdown for all methods corresponding to each subtask of A3CG is found in Table~\ref{tab:full_AAA_ac_ate_results}.

\section{Computation and Tools Used for Our Study}\label{appendix:computational_tools}
This study was conducted with the help of external, publicly available tools (Pytorch, Huggingface, GPT-4o, Llama3, DeepSeek V3, Claude 3.5 Sonnet), with all experiments run on a single NVIDIA GeForce RTX 4090 GPU. We use the T5-base\footnote{https://huggingface.co/google-t5/t5-base} version (for T5, T5+AL, T5+CL) and BERT-base-uncased\footnote{https://huggingface.co/google-bert/bert-base-uncased} version (for BERT-ST, BERT-ST+AL, BERT-ST+CL) from Huggingface\footnote{https://huggingface.co}. GPT-4o is deployed through the OpenAI API\footnote{https://openai.com/index/openai-api/}, Llama3 (70B) through the Groq API\footnote{https://console.groq.com/docs/quickstart}, DeepSeek V3 through the DeepSeek API\footnote{https://platform.deepseek.com/}, Claude 3.5 Sonnet through the Anthropic API\footnote{https://www.anthropic.com/api}. We specify the model sizes in the following (model sizes for proprietary models are unavailable): BERT(110M), T5(220M), Llama3 (70B), DeepSeek V3 (671B), while the model sizes for SOTA-ABSA methods can be found from their respective papers - GRACE~\cite{Luo2020grace}, IT-RER-ABSA~\cite{zheng2024it-rer}, InstructABSA~\cite{scaria2023instructabsa}, CONTRASTE~\cite{mukherjee2023contraste}.

\clearpage
\begin{table*}[h]
    \centering
    \small
    \renewcommand{\arraystretch}{1.3}  % Adjust row spacing
    \setlength{\tabcolsep}{10pt}       % Adjust column spacing
    \begin{tabular}{p{10cm} p{5cm}} 
        \toprule
        \textbf{Sustainability Statement} & \textbf{Labels} \\ 
        \midrule
        We strive to improve patient safety through adequate trainings for our people as well as maintaining open communications with our patients. & (''patient safety'', ``planning'') - aspect category: \textbf{Worker \& Consumer Safety}; (''adequate trainings'', ``planning'') - aspect category: \textbf{Workplace} \\  
        \midrule
        We have also embarked on conservation efforts outside of our oil palm plantations. & (''conservation efforts'', ``implemented'') - aspect category: \textbf{Ecological Conservation} \\ 
        \midrule
        The carbon footprint intensity decreased 5\% from the previous year mainly due to continuous efforts to optimize eco-efficiency in our building operations. & (''carbon footprint intensity'', ``implemented'') - aspect category: \textbf{Emissions Control}; (''eco-efficiency'', ``implemented'') - aspect category: \textbf{Resource Optimization}\\ 
        \midrule
        It is imperative that we manage our business prudently with high standard of corporate governance and integrity. & (''corporate governance'', ``indeterminate'') - aspect category: \textbf{Management} \\
        \midrule
        The Group aims to achieve this by implementing several ethics and  anti-corruption policies across our operations. & (''anti-corruption policies'', ``planning'') - aspect category: \textbf{Business Compliance}\\
        \midrule
        With a PDPA officer appointed, we continue to maintain high standards in safekeeping of personal data of our patients. & (''personal data'', ``implemented'') - aspect category: \textbf{Data \& Cybersecurity Protection}; \\
        \midrule
        Factory closures, as well as labour and capacity shortages have resulted in volatile freight costs and uncertain delivery times. & \textbf{no aspect \& no action} \\
        \bottomrule
    \end{tabular}
    \caption{Samples from the A3CG dataset. Outputs are in the format of (\textbf{aspect}, \textbf{action}) pair(s), and the category of the aspect is also denoted. Pairs, each comprising of an aspect-action pair and aspect category, are delineated by ;}
    \label{tab:dataset_examples}
\end{table*}

\begin{table*}[]
\small
\begin{tabular}{|p{5.5cm}|p{9.5cm}|}
\hline
\textbf{Resource Optimization} & Prioritizes the proactive, efficient, and sustainable selection and use of energy, water, and materials. This approach focuses on minimizing environmental impact through the strategic choice and utilization of resources. Distinct from waste management as it does not primarily deal with waste but rather prevents waste generation by optimizing resource use from the start. \\
\hline
\textbf{Waste Management} & Concentrates on managing waste after it has been produced, including the reduction, handling, disposal, and treatment of solid waste and wastewater. It involves but is not limited to recycling, reusing materials, and employing wastewater treatment processes to mitigate environmental degradation and facilitate resource recovery. \\
\hline
\textbf{Emissions Control} & Targets the reduction of various emissions from company operations, with an emphasis on mitigating climate change impacts or protecting air quality. \\
\hline
\textbf{Ecological Conservation} & Manages the company’s interaction with natural ecosystems, focusing on minimizing harmful impacts and promoting biodiversity and habitat conservation. \\
\hline
\textbf{Workplace} & Focuses on the social aspects of sustainability within the company to promote equity, inclusivity, or quality of life, emphasizing employee well-being, inclusivity, or professional growth.\\
\hline
\textbf{Outreach} & Enhances social sustainability by contributing to community well-being, fostering customer satisfaction, building customer relationships, or collaborating with strategic partners to promote equity, inclusivity, and quality of life. \\
\hline
\textbf{Management} & Ensures strategic, ethical, or effective governance through rigorous oversight, diverse board composition, active stakeholder engagement, or high ethical standards within company leadership. \\
\hline
\textbf{Business Compliance} & Adherence to anti-corruption, anti-competitive practices, financial regulations, or financial reporting. \\
\hline
\textbf{Worker \& Consumer Safety} & Adherence to laws that guarantee the safety of products and services to consumer health, as well as fair labor practices and the protection of workers' rights. \\
\hline
\textbf{Data \& Cybersecurity Protection} & Protection and confidentiality of data, which includes measures against cyber threats and compliance with data security regulations. \\
\hline
% Add other rows as needed
\end{tabular}
\captionof{table}{Sustainability Categories}  % Uses captionof to add a caption outside of a floating environment
\label{tab:taxonomy}
\end{table*}

\clearpage

\begin{table*}[h]
    \small
    \centering
    \renewcommand{\arraystretch}{1.2}
    \begin{tabular}{l p{1.5cm} p{3.2cm} p{3.2cm} p{3.2cm}}
        \toprule
        \textbf{Data Split} & \textbf{Full Dataset} & \textbf{Fold 1} & \textbf{Fold 2} & \textbf{Fold 3} \\
        \midrule
        Seen Categories & 
        All \newline categories
        & 
        Worker \& Consumer Safety, Workplace, Outreach, Management, Business Compliance, Data And Cybersecurity Protection, Ecological Conservation
        & 
        Emissions Control, Resource Optimization, Outreach, Management, Business Compliance, Waste Management
        & 
        Resource Optimization, Worker \& Consumer Safety, Emissions Control, Workplace, Waste Management, Data And Cybersecurity Protection, Ecological Conservation \\
        \midrule
        Unseen Categories & - & 
        Emissions Control, Waste Management, Resource Optimization
        & 
        Worker \& Consumer Safety, Data And Cybersecurity Protection, Workplace, Ecological Conservation
        & 
        Outreach, Management, Business Compliance \\
        \midrule
        Seen Train Samples & 1391 & 850 & 850 & 850 \\
        Seen Val Samples & 214 & 135 & 135 & 135 \\
        Seen Test Samples & 399 & 270 & 270 & 270 \\
        Unseen Test Samples & - & 270 & 270 & 270 \\
        Total Samples & 2004 & 1525 & 1525 & 1525 \\
        \bottomrule
    \end{tabular}
    \caption{Detailed Breakdown of Data Splits. Seen Categories correspond to the categories found within Seen Train, Seen Validation, Seen Test Samples. Unseen Categories correspond to the categories found within Unseen Test Samples. Seen Train, Seen Validation, Seen Test Samples, Total Samples and Unseen Test Samples correspond to the number of samples (i.e. statements) for each data split.}
    \label{tab:data_split_num_samples}
\end{table*}

\begin{table*}[h]
    \small
    \centering
    \renewcommand{\arraystretch}{1.2}
    \setlength{\tabcolsep}{3.8pt} % Moderate column spacing
    \begin{tabular}{p{4.2cm} ccc cccc cccc cccc} % Adjust first column width
        \toprule
        \multirow{2}{*}{\raisebox{-1.3\height}{\parbox{3cm}{\centering \textbf{Data Split/} \\[0.2cm] \textbf{Aspect Category}}}}  & \multicolumn{3}{c}{\textbf{Full Dataset}} & \multicolumn{4}{c}{\textbf{Fold 1}} & \multicolumn{4}{c}{\textbf{Fold 2}} & \multicolumn{4}{c}{\textbf{Fold 3}} \\
        \cmidrule(lr){2-4} \cmidrule(lr){5-8} \cmidrule(lr){9-12} \cmidrule(lr){13-16}
        & \rotatebox{90}{Seen Train} & \rotatebox{90}{Seen Val} & \rotatebox{90}{Seen Test} 
        & \rotatebox{90}{Seen Train} & \rotatebox{90}{Seen Val} & \rotatebox{90}{Seen Test} & \rotatebox{90}{Unseen Test} 
        & \rotatebox{90}{Seen Train} & \rotatebox{90}{Seen Val} & \rotatebox{90}{Seen Test} & \rotatebox{90}{Unseen Test} 
        & \rotatebox{90}{Seen Train} & \rotatebox{90}{Seen Val} & \rotatebox{90}{Seen Test} & \rotatebox{90}{Unseen Test} \\
        \midrule
         Resource Optimisation                & 266 & 33 & 69 & -- & -- & -- & 126 & 199 & 45 & 86 & -- & 238 & 35 & 62 & -- \\
        Management                           & 209 & 25 & 63 & 187 & 26 & 62 & -- & 154 & 29 & 50 & -- & -- & -- & -- & 101 \\
        Workplace                            & 277 & 43 & 81 & 264 & 50 & 66 & -- & -- & -- & -- & 164 & 248 & 19 & 71 & -- \\
        Emissions Control                    & 200 & 34 & 50 & -- & -- & -- & 99 & 175 & 28 & 51 & -- & 165 & 42 & 47 & -- \\
        Worker \& Consumer Safety            & 222 & 35 & 71 & 225 & 28 & 65 & -- & -- & -- & -- & 131 & 193 & 30 & 57 & -- \\
        Business Compliance                  & 112 & 24 & 31 & 126 & 10 & 29 & -- & 79 & 16 & 37 & -- & -- & -- & -- & 67 \\
        Outreach                             & 234 & 35 & 71 & 200 & 38 & 73 & -- & 201 & 23 & 54 & -- & -- & -- & -- & 127 \\
        Waste Management                     & 212 & 29 & 66 & -- & -- & -- & 102 & 187 & 36 & 56 & -- & 180 & 23 & 63 & -- \\
        Data \& Cybersecurity Protection     & 93  & 16 & 20 & 92 & 13 & 24 & -- & -- & -- & -- & 58 & 81 & 12 & 13 & -- \\
        Ecological Conservation              & 64  & 10 & 28 & 64 & 5  & 17 & -- & -- & -- & -- & 27 & 50 & 11 & 16 & -- \\
        \midrule
        Total No Aspect                   & 463 & 73 & 130 & 307 & 47 & 94 & 94 & 297 & 47 & 94 & 94 & 297 & 47 & 94 & 94 \\
        Total Aspect Count$^*$ & 1889 & 284 & 550 & 1158 & 170 & 336 & 327 & 995 & 177 & 334 & 380 & 1155 & 172 & 329 & 295 \\
        \bottomrule
    \end{tabular}
    \caption{Aspect counts for each of the different aspect categories within each split of the dataset. $^*$The total aspect count excludes the count of no aspects. For each data split, the number of no aspect samples are kept at a consistent percentage - around 22\% of the sum of total number of aspects + no aspects.}
    \label{tab:data_split_num_aspects}
\end{table*}

\clearpage

\begin{table*}[h]
    \centering
    \small
    \begin{tabular}{p{3cm} p{12cm}} 
        \toprule
        \textbf{Method(s)} & \textbf{Prompt} \\
        \midrule
        T5, T5+AL, T5+CL & Extract aspect-action pairs from the following sentence: \{\textbf{Sustainability Statement}\} \\
        \midrule
        InstructABSA, IT-RER-ABSA & Definition: The output will be the aspects and the aspect's action classification. In cases where there are no aspects the output should be noaspectterm:none.
        \newline \newline Positive example 1-
        \newline \{\textbf{Few-Shot Example}\}
        \newline \newline Positive example 2-
        \newline \{\textbf{Few-Shot Example}\}
        \newline \newline Now complete the following example-
        \newline Input: \{\textbf{Sustainability Statement}\} \\
        \midrule
        All LLMs (GPT-4o, Claude 3.5 Sonnet, Llama 3 (70B), DeepSeek V3) & 
        [Task Description]
        \newline The aspect-action pair consists of aspect term and action label. You are tasked with extracting aspect-action pairs from sentences.
        \newline
        \newline [Definitions of Aspects]
        \newline Aspects refer to sustainability-related elements found within the sentence that describe entities, goals, sub-areas, activities of sustainability. To enhance specificity, aspects should incorporate appropriate qualifiers that clarify their meaning. At the same time, they should remain succinct by eliminating any redundant language that do not contribute to their core meaning. Aspects must be explicitly found within the sentence and cannot include words not present in the sentence.
        \newline
        \newline [Definitions of Actions]
        \newline Action labels are the type of action taken toward the aspect. Action labels fall under the following categories:
        \newline ``Planning'': Indicates that an action has been planned or a commitment has been made by a company to address or engage with the aspect to advance its sustainability efforts. 
        \newline ``Implemented'': Indicates that an action has already been taken to address or engage with the aspect to advance the company's sustainability efforts. 
        \newline ``Indeterminate'': Indicates that it is unclear from the statement if the company intends to address or engage with the aspect to advance its sustainability efforts, or how it intends to do so.
        \newline
        \newline [Response Instructions]
        \newline Evaluate the explicit content of the sentence to determine the aspect-action pairs. 
        There can be more than one aspect-action pair the sentence. However, each aspect can only have one action. Alternatively, the sentence may not have any aspect-action pairs. In such cases, return (''no aspect'', ``no action'').
        \newline
        \newline [Response format]
        \newline For the sentence, provide the aspect-action pairs in tuples using the following format. Do not output any explanations. You can observe the following example(s) for reference.
        \newline \newline \{\textbf{Few-Shot Examples}\}
        \newline \newline Now complete the following example-
        \newline Input: \{\textbf{Sustainability Statement}\}
        \\
        \bottomrule
    \end{tabular}
    \caption{Prompt templates for all methods that require prompting. ``Sustainability statement'' refers to the dataset samples. For InstructABSA \& IT-RER-ABSA, the prompt is kept similar to the original format of the Sentiment-Pair Extraction (ASPE) task in the respective papers. Few-shot examples for InstructABSA is found in Table~\ref{tab:few_shot_examples_instructabsa}. Few-shot examples for LLMs are found in Table~\ref{tab:few_shot_examples_llm}. Few-shot examples for IT-RER-ABSA are retrieved from the training sets through the retrieval module in its original implementation.}
\label{tab:method_prompts}
\end{table*} 

\begin{table*}[h]
    \centering
    \small
    \begin{tabular}{p{5cm} p{10cm}}
        \toprule
        \textbf{Dataset Partition} & \textbf{Examples} \\
        \midrule
        Full Dataset & Input: For example, in the case of Flu-Pandemic disruption, the SaFe Management Measures System will be activated to prevent pandemic virus transmission at the workplace.
        \newline Output: pandemic virus transmission:planning, workplace:planning  
        \newline \newline Input: We encourage our team to reduce paper usage and to reuse or recycle non-sensitive paper waste where practicable.
        \newline Output: paper usage:indeterminate, non-sensitive paper waste:indeterminate \\
        \midrule
        Fold 1       & Input: For example, in the case of Flu-Pandemic disruption, the SaFe Management Measures System will be activated to prevent pandemic virus transmission at the workplace.
        \newline Output: pandemic virus transmission:planning, workplace:planning 
        \newline \newline Input: Besides scouting for the latest technology and solutions, the company also engages in strategic partnerships with research institutions.
        \newline Output: strategic partnerships:implemented
        \\
        \midrule
        Fold 2       & Input: We reviewed our regulatory risks as part of the management risk assessment process in 2021 and have achieved the 2021 target.
        \newline Output: management risk assessment process:implemented, regulatory risks:implemented
        \newline \newline Input: We strive to adopt best practices to optimize consumption and minimise carbon emissions as an organisation.
        \newline Output: carbon emissions:planning \\
        \midrule
        Fold 3       & Input: We are committed to improve on our occupational health and safety initiatives and conduct regular reviews of our programmes, processes, risk assessments and controls.
        \newline Output: occupational health and safety initiatives:planning, risk assessments and controls:planning
        \newline \newline Input: We encourage our team to reduce paper usage and to reuse or recycle non-sensitive paper waste where practicable.
        \newline Output: paper usage:indeterminate, non-sensitive paper waste:indeterminate \\
        \bottomrule
    \end{tabular}
    \caption{Few-Shot Examples for InstructABSA}
     \label{tab:few_shot_examples_instructabsa}
\end{table*}

\begin{table*}[h]
    \centering
    \small
    \begin{tabular}{p{3.5cm} p{11.5cm}}
        \toprule
        \textbf{Data Partition} & \textbf{Example} \\
        \midrule

        Full Dataset & Input: The Group revitalised and rejuvenated five existing coffeeshops in FY2022 to improve customers' dining experience and hygiene standards. \\
        & Response: (''hygiene standards'', ``implemented''), (''customers' dining experience'', ``implemented'') \\
        & \\
        & Input: It is imperative that we manage our business prudently with high standard of corporate governance and integrity. \\
        & Response: (''corporate governance'', ``indeterminate'') \\
        & \\
        & Input: We are committed to improve on our occupational health and safety initiatives and conduct regular reviews of our programmes, processes, risk assessments and controls. \\
        & Response: (''occupational health and safety initiatives'', ``planning''), (''risk assessments and controls'', ``planning'') \\
        & \\
        & Input: This project will commence in 2022 and take place over the next 3 years. \\
        & Response: (''no aspect'', ``no action'') \\
        & \\
        & Input: In 2020, we achieved a 44\% reduction in carbon emissions intensity against 2007 levels, putting us on track to achieving our SBTi-validated target of 59\% by 2030. \\
        & Response: (''carbon emissions intensity'', ``implemented'') \\
        \midrule

        Fold 1 Seen Test & Input: This project will commence in 2022 and take place over the next 3 years. \\
        & Response: (''no aspect'', ``no action'') \\
        & \\
        & Input: We have in place robust worksite inspection procedures and monthly audits to identify workplace hazards and ensure all activities comply with all Group and regulatory requirements. \\
        & Response: (''workplace hazards'', ``implemented''), (''worksite inspection procedures'', ``implemented'') \\
        & \\
        & Input: Moving ahead, we are targeting to include automation in tracking order status and completion, routing and invoicing to improve customer experience. \\
        & Response: (''customer experience'', ``planning'') \\
        & \\
        & Input: Continuous learning is necessary to help enhance workmen proficiency. \\
        & Response: (''workmen proficiency'', ``indeterminate''), (''continuous learning'', ``indeterminate'') \\
        & \\
        & Input: Biodiversity conservation efforts shaping up across plants. \\
        & Response: (''biodiversity conservation efforts'', ``indeterminate'') \\
        \midrule

        Fold 1 Unseen Test & Input: The upgrading programme of our second generation fleet of 19 C651 trains has commenced and is expected to be completed in 2018. \\
        & Response: (''no aspect'', ``no action'') \\
        & \\
        & Input: In 2020, we achieved a 44\% reduction in carbon emissions intensity against 2007 levels, putting us on track to achieving our SBTi-validated target of 59\% by 2030. \\
        & Response: (''carbon emissions intensity'', ``implemented'') \\
        & \\
        & Input: The carbon footprint intensity decreased 5\% from the previous year mainly due to continuous efforts to optimize eco-efficiency in our building operations. \\
        & Response: (''carbon footprint intensity'', ``implemented''), (''eco-efficiency'', ``implemented'') \\
        & \\
        & Input: In 2019, the Group generated a total of approximately 936,000 tonnes of sludge. \\
        & Response: (''sludge'', ``indeterminate'') \\
        & \\
        & Input: We will continue to maintain or lower our energy and water consumption in FY2022. \\
        & Response: (''energy and water consumption'', ``planning'') \\
        \bottomrule
    \end{tabular}
    \caption{Few-Shot Examples for LLMs - continued on the next page}
     \label{tab:few_shot_examples_llm}
\end{table*}

\begin{table*}[!ht]
    \ContinuedFloat
    \centering
    \small
    \begin{tabular}{p{3.5cm} p{11.5cm}}
        \toprule
        \textbf{Data Partition} & \textbf{Example} \\
        \midrule
        Fold 2 Seen Test & Input: Each goal has specific targets to be achieved over the next 15 years. \\
        & Response: (''no aspect'', ``no action'') \\
        & \\
        & Input: In FY2019, apart from economic performance, we have with the help of an independent external consultant established our sustainability performance management framework. \\
        & Response: (''sustainability performance management framework'', ``implemented'') \\
        & \\
        & Input: We remain committed in ensuring that we handle medical waste in a safe and sustainable manner. \\
        & Response: (''medical waste'', ``planning'') \\
        & \\
        & Input: In 2019, the Group generated a total of approximately 936,000 tonnes of sludge. \\
        & Response: (''sludge'', ``indeterminate'') \\
        & \\
        & Input: Free shuttle bus services are also available at some of its shopping malls. \\
        & Response: (''no aspect'', ``no action'') \\
        \midrule

        Fold 2 Unseen Test & Input: We have in place robust worksite inspection procedures and monthly audits to identify workplace hazards and ensure all activities comply with all Group and regulatory requirements. \\
        & Response: (''workplace hazards'', ``implemented''), (''worksite inspection procedures'', ``implemented'') \\
        & \\
        & Input: This project will commence in 2022 and take place over the next 3 years. \\
        & Response: (''no aspect'', ``no action'') \\
        & \\
        & Input: Continuous learning is necessary to help enhance workmen proficiency. \\
        & Response: (''workmen proficiency'', ``indeterminate''), (''continuous learning'', ``indeterminate'') \\
        & \\
        & Input: We have also embarked on conservation efforts outside of our oil palm plantations. \\
        & Response: (''conservation efforts'', ``implemented'') \\
        & \\
        & Input: We are committed to foster a non-discriminatory workplace environment. \\
        & Response: (''non-discriminatory workplace environment'', ``planning'') \\
        \midrule

        Fold 3 Seen Test & Input: In 2020, we achieved a 44\% reduction in carbon emissions intensity against 2007 levels, putting us on track to achieving our SBTi-validated target of 59\% by 2030. \\
        & Response: (''carbon emissions intensity'', ``implemented'') \\
        & \\
        & Input: We are committed to improve on our occupational health and safety initiatives and conduct regular reviews of our programmes, processes, risk assessments and controls. \\
        & Response: (''occupational health and safety initiatives'', ``planning''), (''risk assessments and controls'', ``planning'') \\
        & \\
        & Input: We encourage our team to reduce paper usage and to reuse or recycle non-sensitive paper waste where practicable. \\
        & Response: (''paper usage'', ``indeterminate''), (''non-sensitive paper waste'', ``indeterminate'') \\
        & \\
        & Input: In order to achieve carbon reduction goals, the governments and relevant regulatory agencies have introduced successively various policies and measures to encourage various industries in implementing green and low-carbon transformation. \\
        & Response: (''carbon reduction goals'', ``indeterminate''), (''green and low-carbon transformation'', ``indeterminate'') \\
        & \\
        & Input: It covers the period from 1 May 2020 to 30 April 2021. \\
        & Response: (''no aspect'', ``no action'') \\

        \bottomrule
    \end{tabular}
    \caption[]{Few-Shot Examples for LLMs - continued on the next page}
\end{table*}

\clearpage
\begin{table*}[!ht]
    \ContinuedFloat
    \centering
    % \vspace{-15cm}
    \small
    \begin{tabular}{p{3.5cm} p{11.5cm}}
        \toprule
        \textbf{Data Partition} & \textbf{Example} \\
        \midrule
        Fold 3 Unseen Test & Input: Each goal has specific targets to be achieved over the next 15 years. \\
        & Response: (''no aspect'', ``no action'') \\
        & \\
        & Input: It is imperative that we manage our business prudently with high standards of corporate governance and integrity. \\
        & Response: (''corporate governance'', ``indeterminate'') \\
        & \\
        & Input: In Singapore, our community programmes and approach are aligned with NCSS' Strategic Thrusts where every person is empowered to live with dignity in a caring and inclusive society. \\
        & Response: (''community programmes'', ``implemented'') \\
        & \\
        & Input: We have also established a business continuity plan (BCP) which focuses on the recovery of technology facilities and platforms, such as critical applications, databases, servers or other required technology infrastructure for the viability of the business. \\
        & Response: (''business continuity plan'', ``implemented'') \\
        & \\
        & Input: Ranked 2nd in customer satisfaction for four consecutive years among supermarket retailers in Singapore. \\
        & Response: (''customer satisfaction'', ``indeterminate'') \\
        \bottomrule
    \end{tabular}
    \caption[]{Few-Shot Examples for LLMs}
\end{table*}

\begin{table*}[h]
    \small
    \centering
    \renewcommand{\arraystretch}{1.2}
    \begin{tabular}{lcccccccc}
        \toprule
        \multirow{2}{*}{$\alpha$} & \multicolumn{4}{c}{T5+AL} & \multicolumn{4}{c}{BERT-ST+AL} \\
        \cmidrule(lr){2-5} \cmidrule(lr){6-9}
        & Fold 1 US & Fold 2 US & Fold 3 US & US Avg & Fold 1 US & Fold 2 US & Fold 3 US & US Avg \\
        \midrule
        0.01 & 45.30  & 44.06 & 39.42 & 42.93 & 26.31 & 21.3  & 29.99 & 25.87 \\
        0.05 & 42.86 & 46.3  & 41.67 & 43.61 & 24.09 & 21.9  & 31.2  & 25.73 \\
        0.1  & 40.27 & 44.82 & 38.77 & 41.29 & 23.28 & 21.49 & 29.31 & 24.69 \\
        0.2  & 36.56 & 47.78 & 41.12 & 41.82 & 26.17 & 21.82 & 28.78 & 25.59 \\
        0.3  & 39.62 & 47.02 & 41.82 & \underline{42.82} & 27.05 & 25.57 & 26.11 & \underline{26.24} \\
        0.4  & 40.88 & 46.27 & 41.05 & 42.73 & 22.63 & 24.12 & 29.31 & 25.35 \\
        0.5  & 40.78 & 43.85 & 34.51 & 39.71 & 27.53 & 25.57 & 29.31 & \textbf{27.47} \\
        0.6  & 41.63 & 45.05 & 42.90  & \textbf{43.19} & 27.53 & 20.65 & 29.31 & 25.83 \\
        0.7  & 42.94 & 44.77 & 35.63 & 41.11 & 27.53 & 25.57 & 29.31 & \textbf{27.47} \\
        0.8  & 36.08 & 45.54 & 37.98 & 39.87 & 27.53 & 22.52 & 29.31 & 26.45 \\
        0.9  & 42.05 & 44.9  & 35.22 & 40.72 & 27.53 & 21.04 & 29.31 & 25.96 \\
        1.0  & 37.45 & 43.39 & 37.39 & 39.41 & 27.53 & 25.57 & 29.31 & \textbf{27.47} \\
        1.1  & 39.39 & 44.69 & 38.7  & 40.93 & 22.63 & 25.57 & 31.74 & 26.65 \\
        1.2  & 42.8  & 45.78 & 35.88 & 41.49 & 27.53 & 25.57 & 29.31 & \textbf{27.47} \\
        1.3  & 40.0  & 45.21 & 36.58 & 40.60 & 27.53 & 25.57 & 29.31 & \textbf{27.47} \\
        1.4  & 37.36 & 45.88 & 36.68 & 39.97 & 27.53 & 25.57 & 29.31 & \textbf{27.47} \\
        1.5  & 42.18 & 45.32 & 40.4  & 42.63 & 27.53 & 25.57 & 28.06 & 27.05 \\
        1.6  & 41.85 & 45.34 & 41.69 & 42.96 & 27.53 & 25.57 & 29.31 & \textbf{27.47} \\
        1.7  & 36.08 & 45.22 & 32.93 & 38.08 & 27.53 & 25.57 & 29.31 & \textbf{27.47} \\
        1.8  & 36.89 & 43.14 & 38.34 & 39.46 & 27.53 & 20.72 & 29.31 & 25.85 \\
        1.9  & 35.13 & 43.14 & 37.37 & 38.55 & 27.53 & 25.57 & 29.31 & \textbf{27.47} \\
        2.0  & 40.87 & 44.14 & 35.56 & 40.19 & 27.53 & 25.57 & 29.31 & \textbf{27.47} \\
        \bottomrule
    \end{tabular}
    \caption{F1 performance of T5+AL and BERT-ST+AL across different $\alpha$ values, for unseen (US) test sets of different folds. US Avg represents average across unseen (US) test sets of all folds. Best US Avg results across all $\alpha$ values are bolded. Results of experimentally selected $\alpha$, 0.3, are underlined.}
    \label{tab:alpha_performance}
\end{table*}

\clearpage

\begin{table*}[h]
    \centering
    \small
    \begin{tabular}{p{2.5cm} p{5.5cm} p{7cm}} 
        \toprule
        \textbf{Error Type} & \textbf{Sample} & \textbf{Analysis} 
        \\ 
        \midrule
        Missing \newline Obvious \newline Aspects & Our innovations also contribute to the reduction of carbon emissions from our delivery fleet due to fewer redeliveries required & When generalizing to an unseen sustainability category, supervised models cannot extract sustainability aspects (i.e. ``carbon emissions'') that are obvious and highly intuitive to sustainability. \\
        \midrule
        i) Elipsis & The company strictly abides by the energy conservation law of the PRC, strives to continuously improve resource efficiency and reduce resource consumption. & All supervised models fail to classify the action associated with the “reduce consumption” aspect as “planning”. This highlights the inability of supervised models to resolve ellipsis. Specifically, “strives to” applies to both “resource efficiency and “resource consumption”, but does not directly precede the latter, therefore requiring contextual inference. 
        \\
        \midrule
        ii) Ambiguous \newline Syntax & During FY2022, we shared a total of S\$129,000 in direct economic value towards community investments, with the decrease in activities related to the pandemic, which restricted face-to-face community interaction. & All supervised models fail to classify the action associated with the “community investments” aspect as “indeterminate.” This highlights their inability to resolve syntactic ambiguity. Specifically, the statement contains an ambiguous modification: it is unclear whether the phrase ``with the decrease in activities related to the pandemic'' modifies ``S\$129,000 in direct economic value towards community investments”, potentially implying a decrease in community investment. In such cases, where ambiguity is present, models fail to classify the action as “indeterminate”. \\ 
        \bottomrule
        
    \end{tabular}
    \caption{Error Analysis for Supervised Models}
    \label{tab:supervised_error_analysis}
\end{table*}

\begin{table*}[h]
    \centering
    \small
        \begin{tabular}{p{2.5cm} p{5.5cm} p{7cm}} 
        \toprule
        \textbf{Error Type} & \textbf{Sample} & \textbf{Analysis} 
        \\ 
        \midrule
        i) Misinterpreting Modality
 & Where possible, we have implemented sustainable measures to monitor our water consumption and increase water efficiency.
 & LLMs fail to classify “indeterminate” as the action associated with the aspects of “water consumption” and “water efficiency” respectively. LLMs fail to recognise that this statement represents uncertainty and tentativeness, misinterpreting the modality involved. Specifically, “where possible” depicts hedging and non-committal language.  
        \\
        \midrule ii) Negation \newline Handling
 & The additional safety measures previously implemented in 2021 also have been reduced or removed accordingly based on each country's local laws and regulations. & LLMs fail to recognise that the “safety measures” have been reduced or removed, demonstrating no enhancement in safety measures, and thereby no improvement in sustainability efforts. By failing to handle this negative case, LLMs fail to classify  “indeterminate” as the action associate with the aspect “safety measures”. \\ 
         \midrule
        iii) Unattributed statements
 & It maps out a framework that a company or organization can follow to set up an effective environmental management system.
 & LLMs fail to classify “indeterminate” for the aspect “environmental management system”. LLMs fail to recognise that this statement is unattributed, meaning that it is unclear who is engaging with the aspect “environmental management system”.  In statements such as these, it is unclear whether the company is directly engages with the aspect, warranting an “indeterminate” classification for the aspect. \\
 \midrule
 iv) Future \newline Dependency
 & When all of the company's Green Mark-awarded projects are fully completed, they will contribute to a reduction in carbon emissions of almost 95,000 tonnes annually.
 & LLMs fail to classify “planning” for the aspect “carbon emissions”. LLMs fail to recognize that reducing carbon emissions is a future-oriented goal that is contingent on a forthcoming event. The goal can only be achieved upon the full completion of the the company's Green Mark-awarded projects, highlighting its forward-looking nature. 
 \\ 
        \bottomrule
    \end{tabular}
    \caption{Error Analysis for LLMs}
    \label{tab:llm_error_analysis}
\end{table*}

\clearpage

\begin{table*}[ht]
    \centering
    \small
    \setlength\tabcolsep{6.3pt} % Adjust column padding
    \renewcommand{\arraystretch}{1.2} % Adjust row spacing
    \begin{tabular}{|c|c|c|cc|cc|cc|c|c|}
        \hline
        \multirow{2}{*}{\textbf{Method}} & \multirow{2}{*}{\textbf{Metric}} & \multirow{2}{*}{\shortstack{\textbf{Full} \\ \textbf{Dataset}}} & \multicolumn{2}{c|}{\textbf{Fold 1}} & \multicolumn{2}{c|}{\textbf{Fold 2}} & \multicolumn{2}{c|}{\textbf{Fold 3}} & \multirow{2}{*}{\textbf{S Avg}} & \multirow{2}{*}{\textbf{US Avg}} \\
        \cline{4-9}
        & & & \textbf{S} & \textbf{US} & \textbf{S} & \textbf{US} & \textbf{S} & \textbf{US} & & \\
        \hline
        \multirow{3}{*}{T5} & ATE & 76.28 & 70.49 & 48.34 & 77.82  & 53.04 & 78.95 & 38.14 & 75.75 & 46.51 \\
                            & AC & 64.19 & 49.85 & 31.00 & 62.52 & 34.72 & 61.30 & 15.51 & 57.89 & 27.08 \\
                            & AAA & 70.48 & 57.85 & 43.03 & 68.90 & 45.74 & 67.94 & 34.59 & 64.90 & 41.12 \\
        \hline
        \multirow{3}{*}{BERT-ST} & ATE & 53.55 & 50.67 & 27.54 & 52.75 & 24.77 & 53.27 & 33.58 & 52.23 & 28.63 \\
                                 & AC & 31.51 & 27.39 & 2.58 & 24.05 & 3.36 & 30.79 & 8.77 & 27.41 & 4.90 \\
                                 & AAA & 43.19 & 39.56 & 25.25 & 39.00 & 22.92 & 43.97 & 30.01 & 40.84 & 26.06 \\
        \hline
        \multirow{3}{*}{T5 + CL} & ATE & 77.05 & 73.38 & 52.65 & 78.54 & 52.48 & 77.80 & 40.92 & 76.57 & 48.68 \\
                                 & AC & 65.72 & 56.80 & 37.02 & 63.19 & 36.87 & 61.40 & 22.34 & 60.46 & 32.08 \\
                                 & AAA & 71.12 & 62.96 & 46.97 & 69.76 & 46.67 & 67.99 & 38.33 & 66.90 & 43.99 \\
        \hline
        \multirow{3}{*}{T5 + AL} & ATE & 75.69 & 71.77 & 43.17 & 77.24 & 54.08 & 76.98 & 45.80 & 75.33 & 47.68 \\
                                 & AC & 64.27 & 53.87 & 26.58 & 59.65 & 37.29 & 58.59 & 29.12 & 57.37 & 31.00 \\
                                 & AAA & 69.27 & 61.24 & 39.62 & 66.91 & 47.02 & 65.17 & 41.82 & 64.44 & 42.82 \\
        \hline
        \multirow{3}{*}{BERT-ST + CL} & ATE & 79.40 & 74.82 & 39.32 & 79.67 & 52.82 & 78.11 & 37.95 & 77.53 & 43.36 \\
                                      & AC & 62.41 & 51.77 & 21.05 & 63.60 & 30.76 & 48.20 & 12.98 & 54.52 & 21.60 \\
                                      & AAA & 68.53 & 60.00 & 37.06 & 69.22 & 41.78 & 58.04 & 34.94 & 62.42 & 37.93 \\
        \hline
        \multirow{3}{*}{BERT-ST + AL} & ATE & 24.30 & 48.77 & 27.63 & 27.13 & 25.57 & 46.51 & 27.60 & 40.80 & 26.93 \\
                                      & AC & 0.00 & 23.75 & 3.19 & 0.00 & 0.00  & 19.28  & 3.79  & 14.34 & 2.33 \\
                                      & AAA & 24.30 & 37.75 & 27.05 & 27.13 & 25.57 & 34.70 & 26.11 & 33.19 & 26.24 \\
        \hline
        \multirow{3}{*}{InstructABSA} & ATE & 75.22 & 70.47 & 40.55  & 74.90 & 55.90 & 73.33 & 51.24 & 72.90 & 49.23 \\
                                      & AC & 65.28 & 53.07 & 23.44  & 57.46 & 40.45 & 58.15 & 38.33 & 56.23 & 34.07 \\
                                      & AAA & 69.47 & 60.23 & 37.53 & 64.73 & 49.76 & 64.14 & 47.38 & 63.03 & 44.89 \\
        \hline
        \multirow{3}{*}{IT-RER-ABSA} & ATE & 76.04 & 71.72  & 65.13  & 76.87  & 55.73  & 77.87 & 41.94 & 75.49 & 54.27 \\
                                     & AC & 63.64 & 50.72 & 30.20 & 59.91  & 39.54 & 63.46 & 24.24 & 58.03 & 31.33 \\
                                     & AAA & 69.20 & 57.70 & 41.81 & 66.02 & 48.87 & 68.83 & 39.10 & 64.18 & 43.26 \\
        \hline
        \multirow{3}{*}{GRACE} & ATE & 75.45 & 74.53 & 59.15 & 76.59 & 51.98 & 76.90 & 50.38 & 76.01 & 53.84 \\
                               & AC & 59.48 & 50.39 & 30.50 & 53.67 & 30.13 & 51.41 & 20.69 & 51.82 & 27.11 \\
                               & AAA & 67.09 & 60.87 & 50.08 & 63.10 & 44.33 & 61.92 & 48.12 & 61.96 & 47.51 \\
        \hline
        \multirow{3}{*}{CONTRASTE} & ATE & 78.32 & 75.43 & 54.55 & 78.02 & 60.07 & 80.61 & 42.03 & 78.02 & 52.22 \\
                                   & AC & 65.14 & 54.28 & 40.69 & 64.61 & 42.45 & 66.46 & 26.46 & 61.78 & 36.53 \\
                                   & AAA & 71.33 & 61.26 & 48.14 & 69.81 & 50.30 & 71.34 & 40.58 & 67.47 & 46.34 \\
        \hline
       \multirow{3}{*}{GPT-4o} & ATE & 42.30 & 40.90 & 52.59 & 49.09 & 41.44 & 46.23 & 39.10 & 45.41 & 44.38 \\
                                            & AC & 26.12 & 24.25 & 37.52 & 32.95 & 25.08 & 30.42 & 20.93 & 29.21 & 27.84 \\
                                            & AAA & 29.79 & 31.61 & 42.98 & 40.00 & 32.51 & 35.58 & 32.35 & 35.73 & 35.95 \\
        \hline
        \multirow{3}{*}{GPT-4o + FS} & ATE & 48.92 & 47.04 & 59.68 & 48.76 & 51.71 & 52.16 & 42.08 & 49.32 & 51.16 \\
                                            & AC & 30.70 & 32.44 & 38.45 & 29.95 & 36.82 & 36.96 & 25.47 & 33.12 & 33.58 \\
                                            & AAA & 35.69 & 39.08 & 46.68 & 39.12 & 41.10 & 40.05 & 33.46 & 39.42 & 40.41 \\
        \hline
        \multirow{3}{*}{Claude 3.5 Sonnet} & ATE & 48.90 & 46.50 & 53.13 & 52.96 & 46.24 & 51.82 & 44.14 & 50.43 & 47.84 \\
                                            & AC & 34.57 & 31.38 & 34.36 & 36.87 & 33.13 & 39.14 & 32.01 & 35.80 & 33.17 \\
                                            & AAA & 37.70 & 36.71 & 39.44 & 41.11 & 36.88 & 41.59 & 38.22 & 39.80 & 38.18 \\
        \hline
        \multirow{3}{*}{Claude 3.5 Sonnet + FS} & ATE & 53.59 & 49.22 & 59.90 & 55.83 & 47.81 & 56.06 & 46.27 & 53.70 & 51.33 \\
                                            & AC & 37.70 & 36.21 & 40.00 & 36.00 & 38.57 & 42.94 & 34.03 & 38.38 & 37.53 \\
                                            & AAA & 42.11 & 40.62 & 46.27 & 43.18 & 40.35 & 45.04 & 39.48 & 42.95 & 42.03 \\
        \hline
        \multirow{3}{*}{Llama 3 (70B)} & ATE & 33.79 & 28.79 & 41.88 & 35.78 & 27.96 & 35.42 & 27.54 & 33.33 & 32.46 \\
                                            & AC & 22.90 & 20.74 & 27.61 & 26.45 & 20.34 & 25.72 & 20.91 & 24.30 & 22.95 \\
                                            & AAA & 20.15 & 17.97 & 25.24 & 23.66 & 18.33 & 22.31 & 18.43 & 21.31 & 20.67 \\
        \hline
        \multirow{3}{*}{Llama 3 (70B) + FS} & ATE & 43.27 & 37.13 & 52.89 & 42.33 & 42.01 & 49.72 & 34.99 & 43.06 & 48.80 \\
                                            & AC & 30.09 & 23.86 & 32.96 & 29.05 & 33.41 &  30.47 & 24.76 & 27.79 & 30.38 \\
                                            & AAA & 29.82 & 25.03 & 38.11 & 33.49 & 33.30 & 33.15 & 26.65 & 30.56 & 32.69 \\
        \hline
        \multirow{3}{*}{DeepSeek V3} & ATE & 47.29 & 45.77 & 57.47 & 52.99 & 47.44 & 51.76 & 41.49 & 50.17 & 48.40 \\
                                            & AC & 33.13 & 27.59 & 41.46 & 36.21 & 33.19 & 36.12 & 23.49 & 33.31 & 32.71 \\
                                            & AAA & 40.16 & 38.06 & 48.18 & 45.27 & 40.23 & 42.46 & 34.84 & 41.96 & 41.08 \\
        \hline
        \multirow{3}{*}{DeepSeek V3 + FS} & ATE & 43.27 & 47.90 & 37.48 & 54.02 & 55.89 & 55.01 & 42.86 & 52.31 & 45.41 \\
                                            & AC & 30.09 & 31.90 & 17.09 & 36.67 & 39.61 & 42.96 & 28.69 & 37.18 & 28.46 \\
                                            & AAA & 35.63 & 38.85 & 30.16 & 44.18 & 44.51 & 46.62 & 34.43 & 43.22 & 36.37 \\
        \hline
    \end{tabular}
    \caption{ATE, AC, and AAA F1 scores, across the full dataset and folds with seen (S) and unseen (US) categories.}
    \label{tab:full_AAA_ac_ate_results}
\end{table*}

\end{document}